\documentclass{article}

\PassOptionsToPackage{numbers, compress}{natbib}
\usepackage{iclr2026_conference, times}
\usepackage{xcolor}
\usepackage[most]{tcolorbox}
\usepackage[framemethod=tikz]{mdframed}
\usepackage{xparse}
\usepackage{tikz}

\newtcbox{\bracketeq}{
  nobeforeafter, enhanced, colback=gray!10, frame empty, math upper,
  boxsep=2pt, left=8pt, right=8pt, top=4pt, bottom=4pt, arc=4pt,
  tcbox raise base, 
  overlay={
    \def\thk{1.0pt}   
    \def\stub{4pt}    
    \draw[line width=\thk] (frame.north west) -- (frame.south west);
    \draw[line width=\thk] (frame.north east) -- (frame.south east);
    \draw[line width=\thk] (frame.north west) -- ([xshift=\stub]frame.north west);
    \draw[line width=\thk] (frame.south west) -- ([xshift=\stub]frame.south west);
    \draw[line width=\thk] (frame.north east) -- ([xshift=-\stub]frame.north east);
    \draw[line width=\thk] (frame.south east) -- ([xshift=-\stub]frame.south east);
  }
}

\newtcbox{\boxeq}{%
  enhanced,
  on line,
  nobeforeafter,
  math upper,          
  frame empty,
  colback=gray!10,
  boxsep=2pt, left=4pt, right=4pt, top=4pt, bottom=4pt,
  arc=3pt,
  tcbox raise base,
  overlay={
    \def\r{3pt}         
    \def\thk{1.0pt}     
    \def\stub{4pt}      
    \def\wipe{1pt}      

    \draw[line width=\thk, rounded corners=\r]
      (frame.north west) rectangle (frame.south east);

    \fill[white]
      ([xshift=\stub,yshift=\wipe]frame.north west)
      rectangle
      ([xshift=-\stub,yshift=-\wipe]frame.north east);
    \fill[white]
      ([xshift=\stub,yshift=\wipe]frame.south west)
      rectangle
      ([xshift=-\stub,yshift=-\wipe]frame.south east);
  }
}
\usepackage{amsthm,amsfonts,amssymb,amsmath}
\usepackage{graphicx}

\usepackage[ruled,vlined]{algorithm2e}
\usepackage{glossaries}
\usepackage{subcaption} 
\usepackage{float}      
\usepackage{comment}
\usepackage{bm}
\usepackage{wrapfig}
\usepackage{enumitem}
\usepackage{subfiles}
\usepackage[export]{adjustbox}
\usepackage[commandnameprefix=always]{changes}
\usepackage[utf8]{inputenc} 
\usepackage[T1]{fontenc}    
\usepackage{hyperref}       
\usepackage{url}            
\usepackage{booktabs}       
\usepackage{amsfonts}       
\usepackage{nicefrac}       
\usepackage{microtype}      
\usepackage{xcolor}         
\usepackage{todonotes}
\usepackage{changes}
\usepackage{cleveref}        
\definecolor{darkblue}{rgb}{0.0,0.0,0.55}

\usepackage[T1]{fontenc}
\usepackage[utf8]{inputenc}

\usepackage{fancyhdr}
\makeatletter
\newcommand{\papertitle}{\@title}
\makeatother

\newtheorem{theorem}{{\bf Theorem}}

\newtheorem{corollary}{{\bf Corollary}}
\newtheorem{definition}{{\bf Definition}}

\newacronym{llm}{LLM}{Large Language Model}
\newacronym{sota}{SOTA}{State-of-the-Art}
\newacronym{sd}{SD}{Speculative Decoding}

\title{The Disparate Impacts of Speculative Decoding}

\author{
  Jameson Sandler\\          
  University of Virginia \\   
  {\small \texttt{jmz4ds@virginia.edu}}
  \And
  Ahmet \"Ust\"un\\          
  Cohere \\   
  {\small \texttt{ahmet@cohere.com}}
  \And
  Marco Romanelli\\          
  Hofstra University \\   
  \small{\texttt{marco.romanelli@hofstra.edu}}
  \AND
  Sara Hooker\\          
  Cohere \\   
  {\small\texttt{sara.hooker@cohere.com}}
  \And
  Ferdinando Fioretto\\          
  University of Virginia \\   
  {\small\texttt{fioretto@virginia.edu}}
}

\iclrfinalcopy 

\begin{document}
\sloppy\allowdisplaybreaks
\maketitle


\fancypagestyle{preprint}{
  \fancyhf{}
  \fancyhead[C]{\small \nouppercase{\papertitle{The Disparite Impacts of Speculative Decoding} — \textit{A Preprint}}}
  \fancyfoot[C]{\thepage}
  \renewcommand{\headrulewidth}{0.3pt}
}

\thispagestyle{plain}   

\pagestyle{preprint}

\begin{abstract}
The practice of speculative decoding, whereby inference is probabilistically supported by a smaller, cheaper, ``drafter'' model, has become a standard technique for systematically reducing the decoding time of large language models. This paper conducts an analysis of speculative decoding through the lens of its potential disparate speed-up rates across tasks.  Crucially, the paper shows that speed-up gained from speculative decoding is not uniformly distributed across tasks, consistently diminishing for under-fit, and often underrepresented tasks.
To better understand this phenomenon, we derive an analysis to quantify this observed ``unfairness'' and draw attention to the factors that motivate such disparate speed-ups to emerge. Further, guided by these insights, the paper proposes a mitigation strategy designed to reduce speed-up disparities and validates the approach across several model pairs, revealing on average a 12\% improvement in our fairness metric.
\end{abstract}

\vspace{-4pt}
\section{Introduction}
\vspace{-4pt}

The rapid growth of large language models (LLMs) has motivated the search for more effective inference paradigms. Among these, speculative decoding \citep{LeviathanKM2023ICML} has emerged as the leading approach for accelerating text generation. This methodology offloads much of the token-generation work onto a lightweight ``drafter'' model, which proposes a set of candidate tokens to be generated. These candidates are then verified by a larger ``verifier'' model in parallel, which accepts or rejects them based on a specific acceptance criteria that can ensure invariance from the vanilla decoding process. When streaks of tokens are accepted by the verifier, an inference speed-up is achieved.
The effectiveness of speculative decoding fundamentally depends on the alignment between the drafter and verifier conditional token distributions. When the drafter and target conditional distributions are well-aligned, acceptance rates are high and throughput gains are substantial. Conversely, misalignment sharply reduces acceptance and erodes speed-up. 

This dependency has motivated a line of work on designing better drafters and verification schemes, e.g., distillation to improve alignment \citep{Zhou_2023_DistillSpec} or structural changes that increase acceptance \citep{Li_2024_EAGLE2}. Yet these advances optimize \emph{average} throughput and say little about how acceleration is distributed across tasks or user groups. 
This gap matters in many application contexts, with particular relevance for multilingual deployment. Multilingual use is a key driver of LLM adoption, but tokenization non-uniformities and data imbalance can conspire to make some languages systematically ``harder'' for both drafting and verification \citep{petrov2023languagemodeltokenizersintroduce}. For example, even when downstream models are identical, subword tokenizers can induce order-of-magnitude differences in sequence lengths across languages, with direct implications for per-request latency and cost budgets \citep{PetrovMTB2023NeurIPS}. 
This raises an important question: \emph{Do certain tasks systematically experience lower speed-ups than others in the context of speculative generation, and can these speed-up disparities be modeled and corrected?} This paper investigates this \emph{computational unfairness} phenomenon and reveals a pattern predictive of systematic slowness: \emph{Tasks to which the drafter exhibits relatively less fitness tend to suffer from lower speed-ups}. This creates a fairness issue where the efficacy of speculative decoding at accelerating inference becomes unevenly distributed across tasks. 



\noindent\textbf{Key contributions.}
To address this phenomenon, this study presents several contributions. First, given next token distributions $p(x), q(x)$, for verifier and drafter models respectively, the paper establishes monotonic links between speculative decoding speed-up, $S$, and different notions of model divergence (i.e., total variation, $\operatorname{TV}(p(x), q(x))$ and cross-entropy, $\operatorname{H}(p(x), q(x))$). Second, based on these results, the paper defines an optimizable and justified notion of speculative decoding unfairness, $\cal U$, built on the smooth divergence function $H(p(x), q(x))$. Next, the paper introduces an analysis that highlights the connections between drafter-fitness and speculative speed-up, establishing disparities in drafter fitness as a predictive source of unfairness. Finally, it showcases the consistent prevalence of speed-up unfairness in a variety of settings and introduces a justified mechanism to mitigate acceleration disparities across tasks, denoted stochastic corrective drafter finetuning (s-CDF).

The results of this work show that speculative decoding could be a potential source of \emph{computational inequity} where some tasks or communities could pay a higher latency to access the same target model. 
We believe that guaranteeing both \emph{accuracy parity} and \emph{acceleration parity} across populations is an important and underexplored objective deserving attention.

\section{Related Work}
\vspace{-4pt}

Speculative decoding has matured from block-wise draft-then-verify ideas into a broad family of methods with theoretical distributional guarantees and practical speed-ups \citep{LeviathanKM2023ICML}. Subsequent work increases acceptance by improving drafter-target alignment, e.g., knowledge distillation and on-policy data \citep{Zhou_2023_DistillSpec}, or by enlarging verified structures (token trees) while keeping outputs faithful to the target model \citep{Li_2024_EAGLE2}. In multilingual inference, specialized drafters can substantially raise acceptance and throughput \citep{YiTHDSY2024EMNLP}. Parallel to these engineering advances, the multilingual tokenization literature documents large cross-language differences in token counts for semantically equivalent inputs, directly affecting runtime and cost \citep{PetrovMTB2023NeurIPS}. These efforts show that acceleration is not merely a property of the \emph{algorithm} but of the \emph{algorithm–population match}. Our work is, to our knowledge, the first to (i) model this interaction explicitly as a fairness question about the \emph{distribution of speed-up} across tasks, and (ii) provide a mitigation that promotes equal acceleration across tasks/languages under constraints on faithfulness. (See Appendix \ref{A:related_work} for extended related work).

\section{Background: Speculative Decoding} \label{sec:background}
\vspace{-4pt}


Let $q(x | s)$ denote the probability distribution induced by the drafter model $Q_{\theta}(s)$ over the token $x$ given context $s$, and let $p(x | s )$ represent the corresponding distribution from the verifier model $P_{\phi}(s)$. We use, $\phi$ and $\theta$ to represent the model parameters, and often suppress the conditioning on $s$ when unambiguous. For a drafted token $x \sim q(x)$, {\bf (1)} if $p(x) \geq q(x)$, the token is immediately accepted. Otherwise, {\bf (2)} the token is rejected with probability $1 - \frac{p(x)}{q(x)}$, and an alternative token is sampled from the residual distribution $p^\prime(x)\triangleq\operatorname{norm}(\max(p(x) - q(x), 0))$. This scheme ensures that generated tokens are sampled from the target distribution $p(x)$~\citep{LeviathanKM2023ICML}, i.e., without loss in generation quality (see Appendix \ref{A:thm3} for proof).

\textbf{Acceptance rate and speed-up.} 
For a given prefix $s$, the (per-step) acceptance rate\footnote{Many references denote the one-step acceptance by \(\beta_{s}\); we use \(\alpha(s)\) for consistency throughout.} $\alpha(s)$ is:
\begin{equation}
    \alpha(s) \triangleq
    \sum_{x\in{\cal V}} q(x | s)    \min \left(1, \frac{p(x| s)}{q(x| s)}\right)
    =
    \sum_{x \in {\cal V}} \min\bigl(q(x| s),\,p(x| s)\bigr),
    \label{eq:alpha}
\end{equation}
where $\cal V$ is the token vocabulary (shared by $P_\phi$ and $Q_\theta$). We then define \(\gamma \in \mathbb N\) to be the number of speculative guesses per iteration (\(\gamma\!=\!1\) is the single-guess case). Let \(c\in[0,1)\) denote the (platform-dependent) drafter cost ratio, {specifically $c = \frac{\text{Drafter Pass Time (s)}}{\text{Verifier Pass Time (s)}}$}.
Under sufficient parallelism to run the \(\gamma+1\) verifier contexts concurrently\footnote{Assuming negligible overhead when $\gamma$ drafted tokens are given to the verifier, then verified in a single pass using parallel inference.}, the expected wall-time improvement factor (vs.\ vanilla decoding) for a given \(s\) is:
\begin{equation}
\mathrm{Speedup}(s;\gamma,c)
\;=\;
\frac{1-\alpha(s)^{\gamma+1}}{(1-\alpha(s))\,[\,\gamma c + 1\,]}.
\label{eq:walltime}
\end{equation}
In particular, for fixed \((\gamma,c)\), Equation~\eqref{eq:walltime} is an increasing function of \(\alpha(s)\). The next section sheds light on consistent speed-up disparities that emerge during the utilization of speculative inference.


\section{Unfairness in Speculative Decoding}
\label{sec:motivation}
With the foundation of speculative decoding established, we now showcase the emergence of speed-up disparities across various languages, motivating this study (and further demonstrated in Section \ref{sec:experiments}).

\begin{wrapfigure}[15]{R}{0.55\textwidth}
  \vspace{-12pt}
  \centering
  \includegraphics[width=\linewidth]{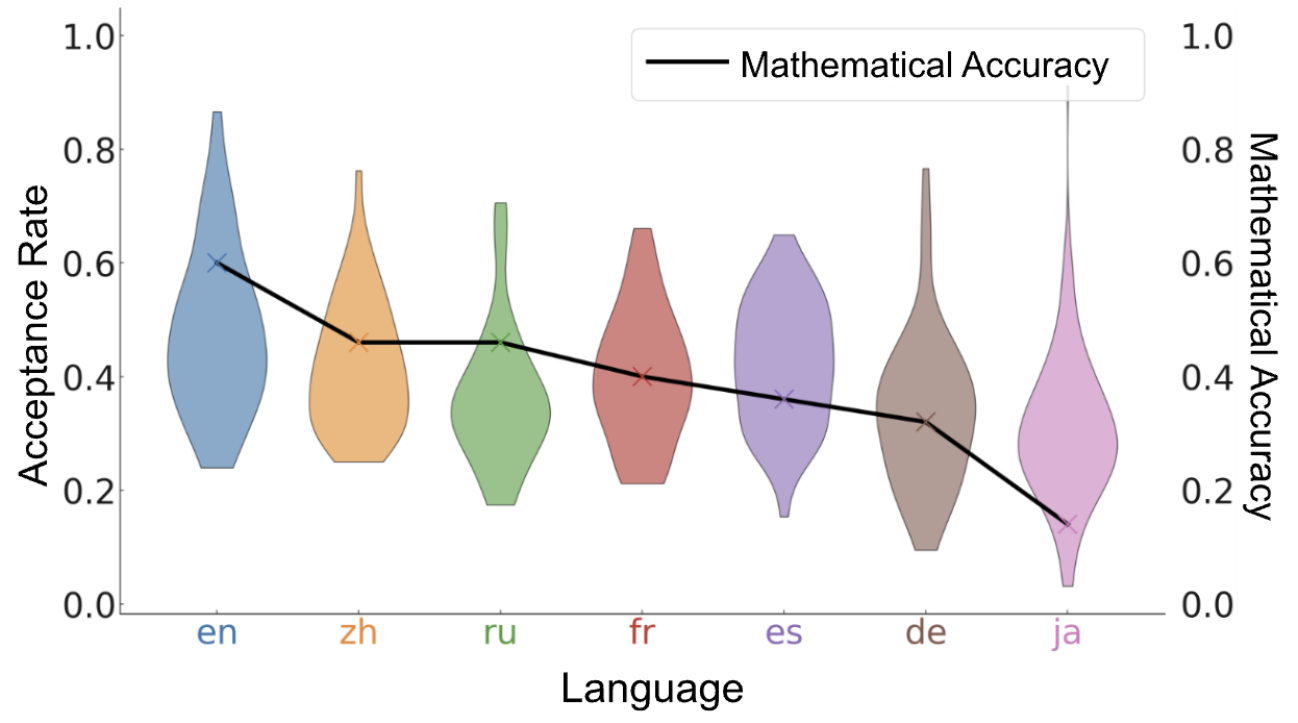}
  \vspace{-13pt}
  \caption{Acceptance rates and accuracies in MGSM using Qwen2.5 series drafter (0.5B) and verifier (3B).} 
  \label{fig:motivation}
\end{wrapfigure}
Figure \ref{fig:motivation} reports the acceptance rates and language-wise benchmark accuracy achieved within speculative decoding on a version of the \emph{multilingual grade-school mathematics} dataset (MGSM) \citep{Shi_2022_MGSM}, evaluating speed-up across various languages. Notably, both accuracy and acceptance rates vary significantly by language, with low accuracy and low speed-up languages coinciding. We find a 13\% gap in average acceptance rates between the fastest and slowest languages, alongside a 52\% difference in accuracy. Japanese, the language with the lowest accuracy, also shows the slowest speed-up. Besides highlighting the existence of speed-up disparities, Figure \ref{fig:motivation} also suggests an important aspect: a connection between {language-wise accuracy} and speed-up as a driver of unfairness, a connection that this paper explores formally and empirically tb sections~\ref{sec:preconditions}, and ~\ref{sec:experiments}.

\vspace{-5pt}
\section{Characterizing Unfairness in Speculative Decoding}
\label{Sec:unfairness}
\vspace{-4pt}

We next discuss
{\bf (i)} why speed-up is monotone in acceptance, 
{\bf (ii)} how speed-up is consequently induced by drafter--verifier \emph{fitness}, and {\bf (iii)} how cross-entropy misfit provides an optimizable surrogate for speed-up unfairness with provable implications for acceleration. All proofs are reported in Appendix \ref{A:proofs}. 

\textbf{Task distributions and task speed-up.}
We start by defining a 'task'. Let \(\mathcal V\) be the shared vocabulary. A \emph{task} \(T\) is a distribution over prefixes \(s\in\mathcal V^{*}\) (e.g., a language or domain). Consider a finite family \(\mathcal T=\{T_1,\dots,T_m\}\); for any task \(T \in {\cal T}\), define the \emph{task-wise acceptance:}
\begin{equation}
    \alpha_T \;\triangleq\; \mathbb E_{s\sim T}\bigl[\alpha(s)\bigr].
\end{equation}
Computing the speed-up for a given task proceeds as follows: Firstly, fix the speculative width \(\gamma \in \mathbb{N}\) and drafter cost ratio \(c\in[0,1)\) as in Section~\ref{sec:background}.
At a prefix \(s\), the expected tokens per verifier iteration is
\(
f_\gamma(\alpha(s))\triangleq \sum_{k=0}^{\gamma}\alpha(s)^{k} = \frac{1-\alpha(s)^{\gamma+1}}{1-\alpha(s)}
\), \citep{Leviathan_2023}, thus the \emph{task-level} speed-up (vs.\ vanilla) on task $T$ is:
\begin{equation}
S_T \triangleq \mathbb E_{s\sim T}[\mathrm{Speedup}(s;\gamma,c)] = \mathbb E_{s\sim T}[\frac{f_\gamma(\alpha(s))}{1+\gamma c}].
\end{equation}


\vspace{-3pt}
\textbf{From divergence to task speed-up.}
Next, we highlight a connection between conventional divergence metrics (i.e., cross-entropy, KL-divergence) and task speed-up $S_T$, which will motivate our ``unfairness'' notion. In particular, the property discussed next makes acceptance the right primitive for analysis and mitigation. 
\begin{theorem} \label{thm:chain}
For \(\gamma\ge 1\), the function \(f_\gamma:[0,1)\to\mathbb R_{+}\), defined as 
\(f_\gamma(\alpha)=\sum_{k=0}^{\gamma}\alpha^{k}\), is increasing on \([0,1)\) and convex for \(\gamma\ge 2\) (linear when \(\gamma=1\)).
Consequently, for fixed \((\gamma,c)\), the task-level speedup on task T can be lower-bounded by the following divergence functions, resulting in the chain:
\begin{equation}
\boxeq{
S_T
\;\ge\; \frac{f_\gamma(\alpha_T)}{1+\gamma c}
\;\ge\;
\frac{f_\gamma\!\Big(1-\sqrt{\tfrac12\,\mathrm{KL}_T}\Big)}{1+\gamma c}
\;\ge\;
\frac{f_\gamma\!\Big(1-\sqrt{\tfrac12\,\bm{D}_T}\Big)}{1+\gamma c}},
\end{equation}
\end{theorem}
where $\mathrm{KL}_T$ is the task-wise Kullback–Leibler divergence: 
$\mathrm{KL}_T\triangleq \mathbb E_{s\sim T}[D_{\mathrm{KL}}(p(\cdot\mid s)\,\Vert\,q(\cdot\mid s))]$, and $\bm{D}_T$ is the task-wise cross-entropy: 
$\bm{D}_T \triangleq \mathbb E_{s\sim T}[-\sum_x p(x|s)\log q(x|s)]$ (proven in Appendix \ref{A:tm1}). 

The result above highlights three key messages: First, 
it places \(\alpha\) as the singular determinant of speculative speed-up at fixed \((\gamma,c)\), 
second, the task-level speedup $S_T$ is increasing in $\alpha_T$ (further corollary in Appendix \ref{A:corr1}),
and third, and most importantly, task-level speed is \emph{monotone} in \(\bm{D}_T\); In particular, the above yields a monotone chain
\(\bm{D}_T \downarrow \;\Rightarrow\; \alpha_T \uparrow \;\Rightarrow\; S_T \uparrow\) \footnote{Throughout, we write $\uparrow x$ ($\downarrow x$) to represent an increase (decrease) to a scalar $x$.}.
This has an important consequence: for fixed \((\gamma,c)\), \(\bm{D}_T\) provides an \emph{optimizable} metric whose reduction \emph{monotonically} tightens a task speed-up. This rationale makes $\bm{D}_t$ a fitting choice for computing unfairness.


\textbf{Unfairness as divergence dispersions.} Next, the section introduces a fairness notion built on the divergence $\bm{D}_T$, from the rightmost expression of Theorem \ref{thm:chain}. 
\begin{definition}
    (Speculative Decoding Unfairness) For a task family ${\cal T}=\{T_i\}_{i=1}^{m}$, \emph{speculative decoding unfairness} is defined as:
    \begin{equation} \boxeq{
        \mathcal U(\mathcal T)
        \;\triangleq\;
        \frac{1}{m}\sum_{T \in \cal T}\bigl(\bm{D}_T - \bm{D}_{\min}\bigr)^2,
        \quad \text{where} \quad
        \bm{D}_{\min}\;\triangleq\;\min_{T\in \cal T}\bm{D}_T}.
        \label{eq:surrogate}
    \end{equation}
\end{definition}
The larger the quantity ${\cal U}({\cal T})$ is, the more disparate the speedups across tasks.  
Note also that reducing \(\mathcal U(\mathcal T)\) contracts the spread of the lower bounds \(\{g(\bm{D}_T)\}_{T \in \cal T}\), where \(g(d)\triangleq f_\gamma\!\big(1-\sqrt{\tfrac12\,d}\big)/(1+\gamma c)\) is decreasing. Therefore, by Theorem \ref{thm:chain}, $\downarrow{\cal U}({\cal T})$ contracts a certified lower envelope of \(\{S_T\}_{T \in \cal T}\).

\section{Preconditions For Speed-Up Disparities}\label{sec:preconditions}
\vspace{-4pt}

So far, we have established that speed-up disparities appear across tasks, and have formalized \emph{speed-up unfairness} via acceptance (and cross-entropy) misalignment. We now ask: \emph{why do these disparities arise so persistently?} Our thesis is that \emph{disparities in acceleration are primarily driven by disparities in \emph{drafter fitness} across tasks}. We make this precise by relating acceptance to model-task alignment.

\textbf{Task misalignment and task fitness.}
We first define what is meant by \emph{task-fitness}, and use these notions to reason about the factors that influence task speed-up. For a given task \(T\) (a distribution over prefixes \(s\in\mathcal V^{*}\)), let
\(u(\cdot\mid s)\) denote the latent task posterior over next tokens
(the conditional distribution that generates the data on task \(T\)).
We define \emph{task misalignments} as follows:
\begin{equation}
    r_p\;\triangleq\;\mathbb{E}_{s \sim T}\big[\tfrac12\!\sum_{x\in\mathcal V}\!\big|u(x\mid s)-p(x\mid s)\big|\big],
    \qquad
    r_q\;\triangleq\;\mathbb{E}_{s \sim T}\big[\tfrac12\!\sum_{x\in\mathcal V}\!\big|u(x\mid s)-q(x\mid s)\big|\big].
\end{equation}
Intuitively, $r_p$ (resp. $r_q$) is the average total variation distance between the verifier’s (resp. drafter’s) next-token distribution and the task’s true posterior  (i.e., the expected fraction of probability mass on which the models disagree with $T$ for some prefix, $s$) thus $1-r$ quantifies 'model$\rightarrow$task' alignment. We therefore interpret \(1-r_p\) and \(1-r_q\) as the \emph{task fitness} of the verifier and drafter, respectively.
The next result formalizes a relation between this drafter task fitness ($1-r_q$) and associated task acceptance.

\begin{theorem}[Drafter-fitness estimator for acceptance]
\label{thm:alpha_est}
Assume $r_p \leq r_q$,\footnote{This assumption is typically trivially satisfied: Running speculative decoding in cases where the drafter preforms better on tasks than the verifier means that generation becomes slower, and of poorer quality relative to vanilla decoding with the drafter alone.} then for any task \(T\):
\begin{equation}
\big|\alpha_T - (1-r_q)\big| \;\le\; r_p,
\end{equation}
\end{theorem}
(proven in \ref{A:thm2}). In particular, when the verifier is well-fit to $T$ (\(r_p\) small), the acceptance \(\alpha_T\) is tightly approximated by the drafter fitness \(1-r_q\). 
Theorem~\ref{thm:alpha_est} formalizes the operational intuition: \emph{once the verifier is reasonably aligned with the task, the drafter’s task fitness becomes the principal driver of acceptance (and hence acceleration).}
Combining Theorems ~\ref{thm:chain} and ~\ref{thm:alpha_est}reveals the monotone chain:
\[
\text{drafter fitness } (1-r_q)\;\uparrow \;\Longrightarrow\; \alpha_T \;\uparrow \;\Longrightarrow\; \text{speed-up} (S_T) \;\uparrow.
\]
The \emph{fairness} implication of this chain is that \emph{given a verifier with high task-fitness, tasks with low drafter fitness will tend to be slower}. 
In other words, Theorem \ref{thm:alpha_est} implies that \emph{disparities in drafter task-fitness have a tendency to produce disparities in speed-up}, with under-fit (and perhaps under-represented) tasks consistently receiving less boost. Accordingly, corollary in Appendix \ref{A:corr2} reveals sufficient conditions for task disparities.

\begin{figure}[t]
  \centering
  \begin{subfigure}[t]{0.4\textwidth}
    \centering
    \includegraphics[width=\linewidth]{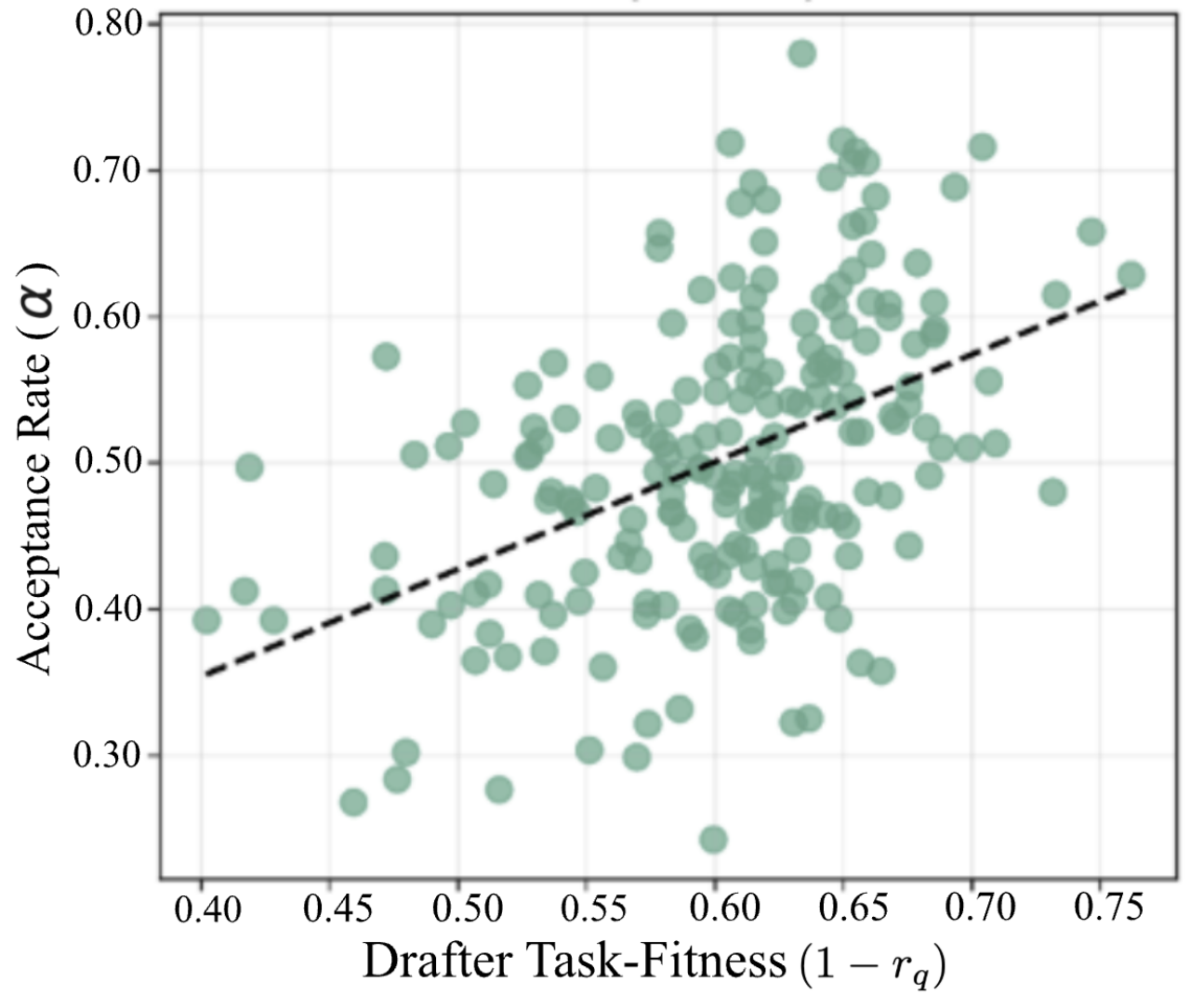}
    \vspace{-5pt}
    \caption{MGSM over Qwen2.5-0.5B and Qwen2.5-3B models}
    \label{fig:ahat-a-1}
  \end{subfigure}\hfill
  \begin{subfigure}[t]{0.55\textwidth}
    \centering
    \includegraphics[width=\linewidth]{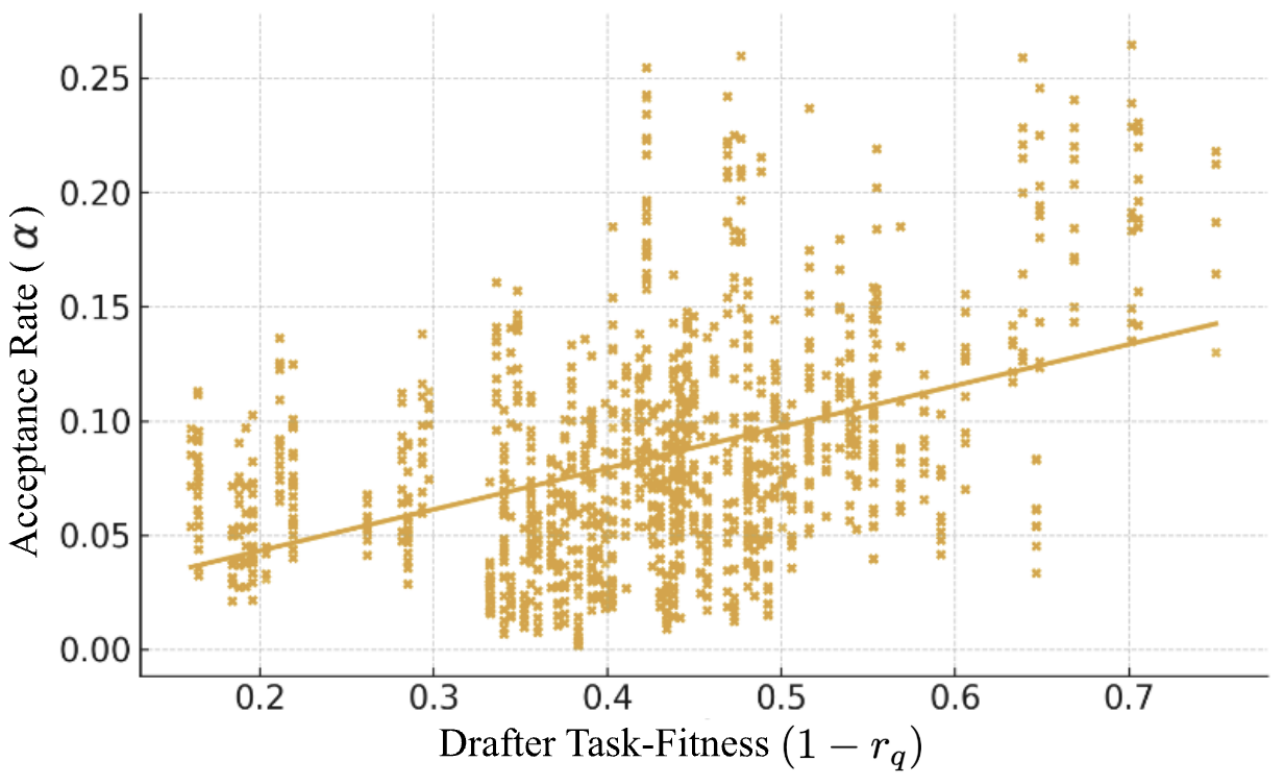}
    \vspace{-5pt}
    \caption{MCoT over Qwen2.5-(0.5B-1.5B), (0.5B-3B), (1.5B-7B), (1.5B-14B), (0.5B-14B) model pairs.}
    \label{fig:ahat-a-2}
  \end{subfigure}\hfill
  \vspace{-5pt}
  \caption{Relation between drafter task-fitness $(1-r_q)$ and task speed-up / acceptance rates $\alpha$.}
  \label{fig:a_hat_sec}
  \vspace{-8pt}
\end{figure}

Figure \ref{fig:a_hat_sec} provides empirical evidence from our experiments that evaluate this dependence. It reveals a strong correlation between drafter fitness and speed-up. Firstly, evaluating acceptance rates and drafter fitness over individual examples from the smaller MGSM \citep{shi2022language}, (with respect to the $P_\phi=$Qwen2.5-3B, $Q_\theta=$0.5B model pair), reveals a Spearman coefficient of $r=0.44$ between drafter fitness and $\alpha$, shown in Figure \ref{fig:ahat-a-1}. We further evaluate the \emph{fitness-speed} relationship on the MCoT \citep{lai2024mcotmultilingualinstructiontuning} dataset, leveraging a large set of model pairs ranging in size from 0.5B, to 14B parameters, and aggregating the associated metrics, Figure \ref{fig:ahat-a-2}. We see once again (especially at the extremes), that large drafter task fitness is associated with larger speed-ups and vice versa. These results provide a clear indication that \emph{under-fit tasks are disadvantaged by disparate slowness} (further results in support of this point are provided in Section \ref{sec:experiments}).

\vspace{-10pt}
\section{Unfairness Mitigation}
\vspace{-4pt}

Motivated by the outlined observations, this section proposes a procedure to reduce speed-up disparities by updating exclusively the \emph{drafter} parameters \(\theta\) while keeping the \emph{verifier} \(P_\phi\) fixed (to preserve the native decoding behavior of the target model). Indeed, by Theorem \ref{thm:chain}, lowering \(\bm D_T\) increases a certified lower bound on acceptance \(\alpha_T\) and hence raises speed-up \(S_T\) monotonically.
Adjusting \(P_\phi\) would compromise exactness and downstream behavior.

\textbf{A fairness-weighted descent direction.}
To prioritize slow tasks (large $\bm{D}_T$ while avoiding any incentive to \emph{increase} divergence on the faster tasks, we propose to 
scale each task’s gradient by its \emph{excess divergence}:
\begin{equation} 
\vspace{-4pt}
\boxeq{
\widehat{\nabla}_\theta\, \mathcal U
\;\triangleq\;
\frac{1}{m}\sum_{T \in \cal T}\bigl(\bm D_T - \bm D_{\min}\bigr)\,\nabla_\theta \bm D_T.}
\label{eq:proj_grad_constr}
\vspace{-0pt}
\end{equation}
Equation ~\eqref{eq:proj_grad_constr} is exactly the gradient of the objective
\(
\widehat{\mathcal U}({\cal T})
= \frac{1}{m}\sum_{T \in \cal T} (\bm D_T -c)^2\), where $\bm{D}_{\min}$ is treated as a constant $c$. This has two immediate benefits:
{\bf (i)} it pushes down on tasks in proportion to how unfairly slow they are, and
{\bf (ii)} the best task ($\bm{D}_T = \bm{D}_{\min}$) receives zero weight, so there is \emph{no explicit term} that increases its divergence. 
Equation \eqref{eq:proj_grad_constr} performs variance-reduction of $\{\bm D_T\}_{T \in \cal T}$ around the current floor while still monotonically decreasing the mean divergence.

Now note that, assume a unique minimizer \(\bm{D}_{\min}=\min_T \bm D_T\).
Away from ties, differentiating \(\mathcal U\) yields:
\(
\nabla_\theta \mathcal U
= \frac{2}{m}\!\left[ \sum_{T \in \cal T}(\bm D_T-\bm D_{\min})\,\nabla_\theta \bm D_T
\;-\; \textcolor{darkblue}{\sum_{T \in \cal T}(\bm D_T-\bm D_{\min})\nabla_\theta \bm D_{\min}}\right].
\)
The second term (blue colored) is problematic: it can \emph{intentionally} move \(\bm D_{\min}\) upward to reduce dispersion, thereby degrading speed-up on the best task.
The gradient proposed in Equation~\eqref{eq:proj_grad_constr} removes this term and thus avoids any \emph{direct} incentive to harm \(\bm D_{\min}\).

\textbf{Stochastic corrective drafter fine-tuning (s-CDF).}
In practice, Equation \ref{eq:proj_grad_constr} is applied by estimating \(\bm D_T\) and \(\nabla_\theta \bm D_T\) from mini-batches. For a batch \(\mathcal B_T\subset T\):
\begin{equation}
\widehat{\bm D}_T
= \frac{1}{|\mathcal B_T|}\sum_{s\in\mathcal B_T}\!\Big[-\!\sum_x p(x\mid s)\log q_\theta(x\mid s)\Big],
\quad
\nabla_\theta \widehat{\bm D}_T
= -\,\mathbb E_{s\in \mathcal B_T,\,x\sim p(\cdot\mid s)}\!\big[\nabla_\theta \log q_\theta(x\mid s)\big].
\vspace{-0pt}
\end{equation}
We adopt this batched approach for temporal efficiency, enabling fast computation of task-wise disparities, and then apply the gradient in Equation \ref{eq:proj_grad_constr} per step resulting in the corrective finetuning process defined in Algorithm \ref{alg:corrective_drafter}.

\begin{wrapfigure}[10]{R}{.48\textwidth}
\vspace{-58pt}
\centering
\begin{algorithm}[H]
\caption{Stochastic Corrective Drafter Fine-tuning (s-CDF)}
\label{alg:corrective_drafter}
\KwIn{\(\{T_1,\dots,T_m\}\); verifier \(P_\phi\) (frozen); drafter \(Q_\theta\); optimizer \(A\); batch sizes \(\{\mathcal B_T\}_{T \in \cal T}\); step size \(\beta\).}
\While{not converged}{
  Sample mini-batches \(\mathcal B_T \subset T\) of size \(\mathcal B_T\) for all \(T \in \cal T\).
  \\
  Estimate \(\widehat{\bm D}_T,\,\nabla_\theta \widehat{\bm D}_T\) on each \(\mathcal B_T\).
  \\
  Set \(\bm{D}_{\min} \leftarrow \min_T \widehat{\bm D}_T\) and \(c \leftarrow \widehat{\bm D}_{\min}\).
  \\
  \(\Delta_\theta \leftarrow -\,\frac{1}{m}\sum_{T \in \cal T}(\widehat{\bm D}_T-c)\,\nabla_\theta \widehat{\bm D}_T\).
  \\
  \(\theta \leftarrow A(\theta,\,\beta,\,\Delta_\theta)\).
}
\end{algorithm}
\end{wrapfigure}

\vspace{-5pt}
\section{Experimental Analysis} 
\label{sec:experiments}
\vspace{-4pt}

In this section, building upon and extending the theoretical insights and fairness objectives discussed, we present key findings from our empirical analysis. The analysis will highlight that: {\bf (1)} the computational speed-up gained from speculative decoding is not uniformly distributed, (validated across multiple model pairs and datasets). {\bf (2) } Languages that receive disproportionate speed-up tend to be underrepresented within conventional training corpus's.  {\bf (3)} Stochastic corrective drafter finetuning serves as an effective speed-up unfairness mitigation across multiple models.

\vspace{-4pt}
\subsection{Datasets And Models}
\vspace{-4pt}

We model each task as a distinct linguistic group and conduct experiments over seven model pairs from the Qwen2.5 family, and four multilingual datasets, spanning bilingual pretraining, multilingual open-ended generations and multilingual mathematics:\\[4pt]
$\bullet$ \textbf{Bilingual Web-Text:}
We use English (L1) and Japanese (L2) web text corpora as seed-text, as over- and under-represented languages, respectively. The English dataset is sourced from a small web-text corpus \citep{nampdn-ai_tiny-webtext_2023}, and Japanese data is drawn from the Japanese WikiNews and related sources curated in \citep{fujiki_llm-japanese-dataset_wikinews_2023}.\\ 
$\bullet$ \textbf{Multilingual Dolly \citep{ustun2024multidolly200}:}
This dataset is drawn from the Aya Evaluation Suite \citep{singh-etal-2024-aya} containing \textbf{open-ended instruction-following} examples. We use its machine-translated version, which includes 200 aligned prompt-response pairs per language. Languages are selected from the set officially supported by the Qwen2.5 series, and their relative representation is estimated using internal linguistic prior probabilities\footnote{We estimate representation based on the prior distribution of drafter/verifier models, detailed in Section 8.2.}. We use BLEU \citep{papineni-etal-2002-bleu} as a reference-based metric for assessing the fidelity of our model outputs relative to reference solutions. \\ 
$\bullet$ \textbf{MGSM (Multilingual Grade-School Math) \citep{shi2022language,lai2024mcotmultilingualinstructiontuning}:} 
This benchmark consists of {\bf grade-school mathematical problems}. It is used to examine the relations between task accuracy and speed-up ($\alpha$), as well as to evaluate the efficacy of our s-CDF method. Given the sample intensive nature of finetuning we use the larger MCoT \citep{lai2024mcotmultilingualinstructiontuning} version of the MGSM for s-CDF experiments, as well as the smaller MGSM \citep{shi2022language} for evaluation-heavy experiments. The benchmark provides reference solutions that can be used to verify model accuracy in a discrete manner.

Additionally, we evaluate seven pairs of models in total. For the bilingual experiments, we use \textbf{GPT-2} (117M) \citep{openai_gpt2_2019} as the drafter and \textbf{GPT-2-XL} (1.5B) \citep{openai_gpt2xl_2019} as the verifier, both of which were pretrained on a private web-text based corpus. For Multilingual Dolly we focus on the \textbf{Qwen2.5-0.5B} and \textbf{Qwen2.5-3B} pair. Similarly for MGSM, and s-CDF experiments we use five pairs from the \textbf{Qwen-2.5} family, with sizes ranging from 0.5B to 14B parameters.\footnote{We opt for Qwen2.5 due to their broad multilingual support and the variety of model sizes that are offered.}

\vspace{-4pt}
\subsection{Disparate Speed-Ups and Underrepresented Languages}
\vspace{-4pt}

We now extend the results presented in Section \ref{sec:motivation}, showing that disparities in speculative decoding speed-ups are consistent, as well as provide evidence in favor of relationships between language ``representation'' and speed-up. 

\begin{wrapfigure}[15]{R}{0.40\textwidth}
  \centering
  \includegraphics[width=\linewidth]{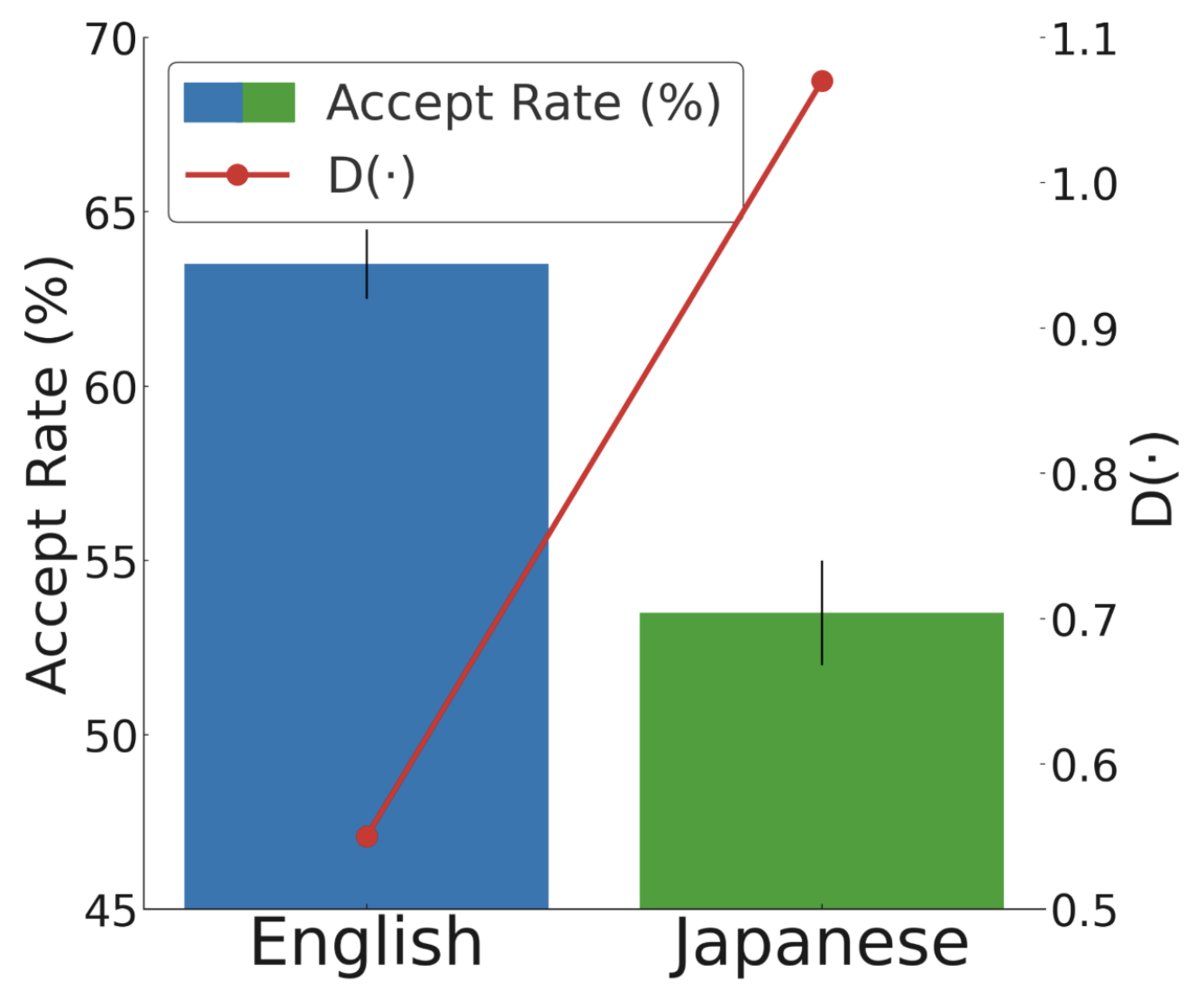}
  \vspace{-16pt}
  \caption{Divergence $\bm{D}(\cdot)$ is high (low) for slower (faster) language.}
  \label{fig:3}
\end{wrapfigure}
\textbf{Bilingual disparities.} We begin by comparing speculative decoding performance between English and Japanese. As shown in Figure~\ref{fig:3}, the acceptance rate $\alpha$ is consistently higher for English than for Japanese (62.5\% vs. 54.5\%). This mirrors the misalignment levels between drafter and verifier on each language, with divergence $\bm{D}(\cdot)$ measured at 0.47 for English and 1.08 for Japanese. This bilingual setup highlights clear speed-up disparity. 

\begin{wrapfigure}[13]{R}{0.45\textwidth}
  \vspace{-12pt}
  \centering
  \includegraphics[width=\linewidth]{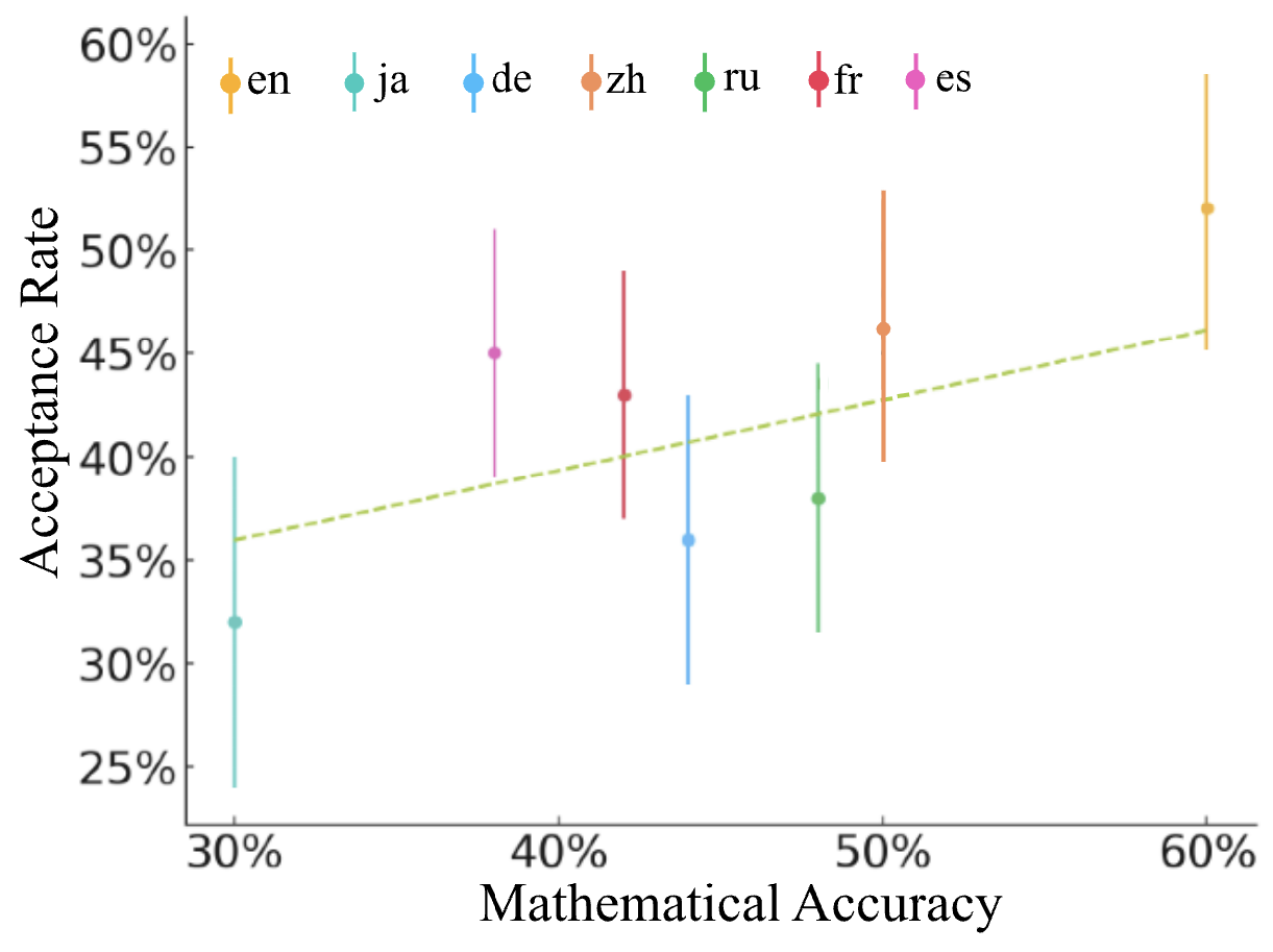}
  \vspace{-16pt}
  \caption{Alpha against task accuracy within different languages on MGSM data.}
 \label{fig:4}
\end{wrapfigure}

\textbf{Multilingual grade-school math (MGSM).} 
We further evaluate this phenomenon in the context of mathematical reasoning using the MGSM benchmark. For each supported language, speculative decoding is performed over chain-of-thought style problem-solving prompts. Final answers are evaluated via pattern matching to compute task accuracy, and speculative decoding is run with Qwen2.5-0.5B, Qwen2.5-3B, for drafter and verifier respectively.  Ultimately, Figure~\ref{fig:4} reveals a strong positive correlation between per-language accuracy and acceptance rate $\alpha$, indicating that \emph{languages benefiting from higher fitness (higher resource languages) also benefit from lower latency during decoding}. 

\begin{wrapfigure}[15]{R}{0.45\textwidth}
  \vspace{-1pt}
  \centering
  \includegraphics[width=\linewidth]{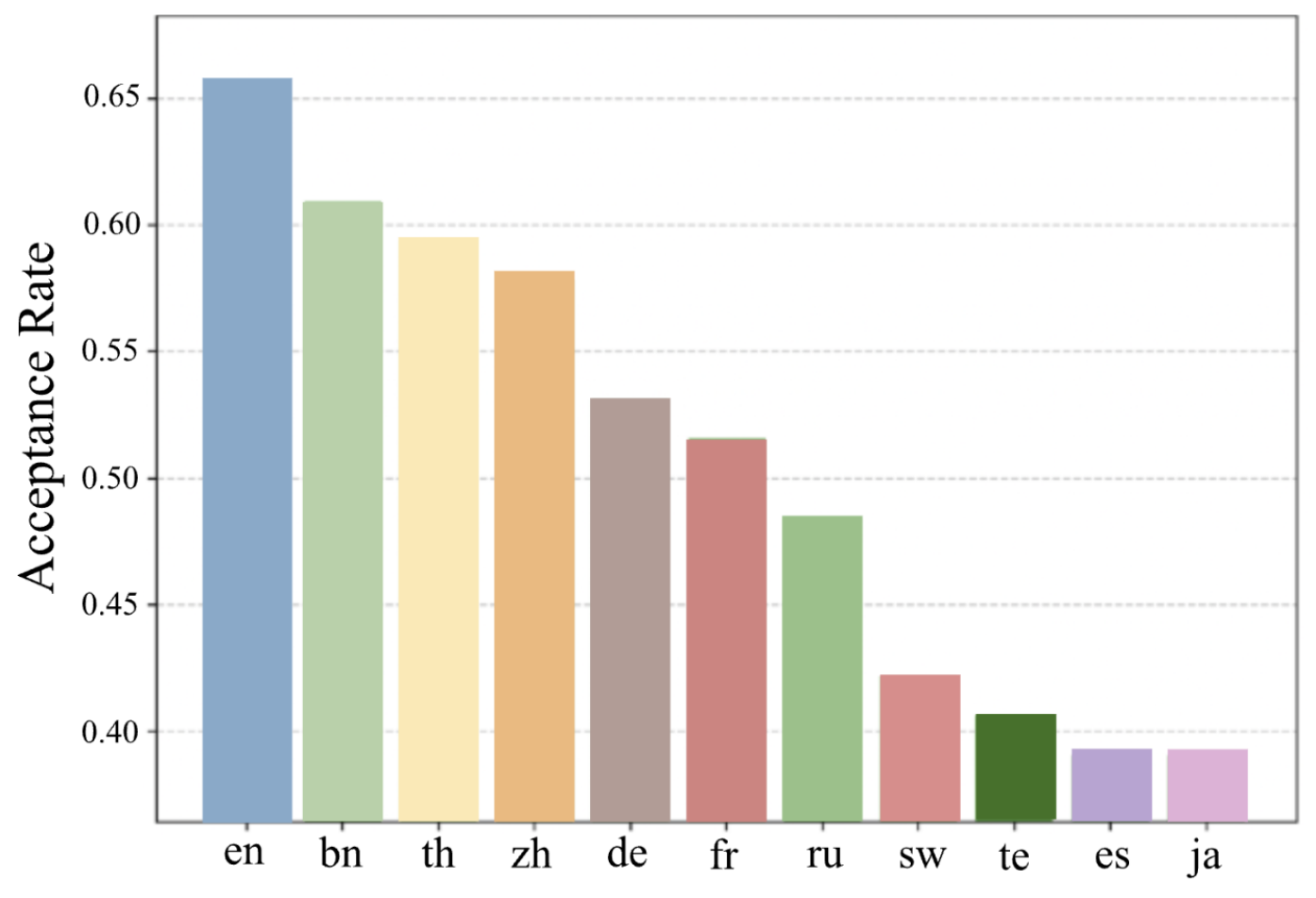}
   \caption{Acceptance rates for Qwen2.5-0.5B, 3B model pair, on larger MCoT dataset.} 
   \label{fig:5}
   \vspace{-5pt}
\end{wrapfigure}

We also explore the larger MGSM benchmark, (MCoT) \citep{lai2024mcotmultilingualinstructiontuning}, featuring an extensive number of languages and mathematical questions, including rare, low-resource languages such as Bangla and Telugu. We utilize the same model pair; drafter (Qwen2.5-0.5B) and verifier (Qwen2.5-3B), then evaluate  acceptances rates over problems on each language, reported in Figure \ref{fig:5}. We continue to see large disparities of up to 65\% between our fastest and slowest languages (English vs Japanese), with languages like Telugu experiencing 60\% lower speed-up relative to English, revealing that \emph{large speed-up disparities persist across variations in dataset}, as well as across different groups of languages.

\textbf{Multilingual DollyQA.}
Next, we evaluate $\alpha$ across a diverse set of languages supported by both DollyQA and Qwen 2.5. For each language $L$, we randomly sample prompts and compute the average acceptance rate $\alpha_L$ under speculative decoding.  

To investigate the relationship between $\alpha_L$ and language representation, we first sample $K$ generations from our drafter on an empty prefix $\emptyset$\footnote{Specifically, we pass the empty string ' ' to our models.}, resulting in the collection of generations: $X = (s_1\sim Q_\theta(\emptyset), \dots, s_K\sim Q_\theta(\emptyset))$. We then confirm the language of each sample with an operator $F(s)$\footnote{Using the NLTK library to classify languages \citep{loper2002nltknaturallanguagetoolkit}.}, resulting in a set of corresponding languages $L = (l_1=F(s_1), \dots, l_K=F(s_K))$. We then estimate the vector $P$, where $P_i = p(l_{i})$ denotes the probability of sampling from language $l_i$ for $N$ valid languages. We compute this estimate with $P \approx \hat P = \frac{1}{K}(\Sigma_{s_j \sim X} \mathbb{I}[(F(s_j )=l_1], \dots, \Sigma_{s_j \sim X} \mathbb{I}[(F(s_j )=l_N])$\footnote{Approximation is shown due to finite sampling error. In practice we set a large $K$ of 100,000.}. The resulting probabilities, $\hat P$ represent model priors with respect to our languages. We finally rank our languages by their estimated representation probability from 'most' to 'least' represented. 

\begin{wrapfigure}[12]{R}{0.47\textwidth}
  \vspace{-18pt}
  \centering
  \includegraphics[width=\linewidth]{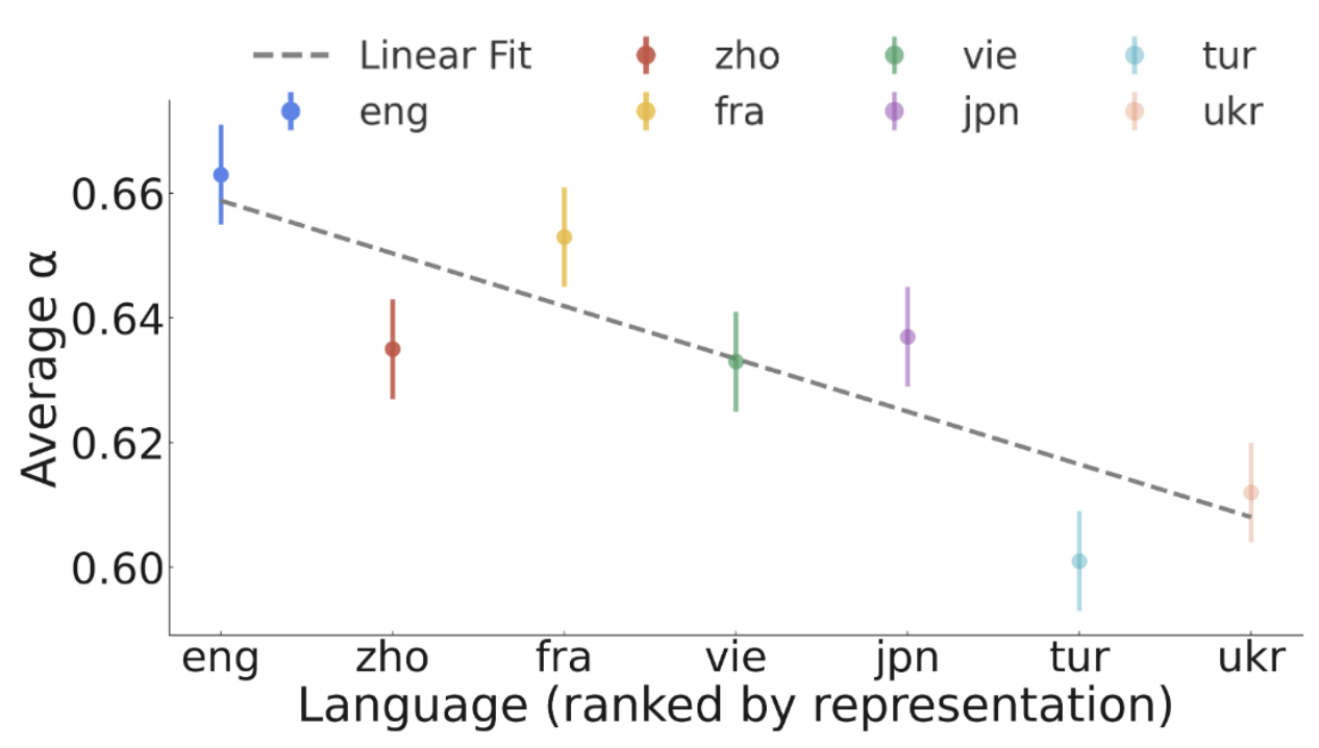}
  \caption{Expected speed up for each language sorted by \emph{representation rank}.}
  \label{fig:6}
\end{wrapfigure}

The resulting trend is clear. As illustrated in Figure~\ref{fig:6} (next page), we observe a strong inverse correlation between language rank and $\alpha$, indicating that less represented languages tend to exhibit lower speed-ups, further demonstrating the disproportionate speed-up benefit high-resource languages receive during speculative-decoding.

\vspace{-4pt}
\subsection{Mitigating Solutions}
\vspace{-4pt}
Next, we assess multiple strategies for reducing disparities in speculative decoding, establishing s-CDF as a promising mitigation method.

\begin{figure}[h]
  \centering
  \begin{subfigure}[t]{0.44\textwidth}
    \centering
    \includegraphics[width=\linewidth,]{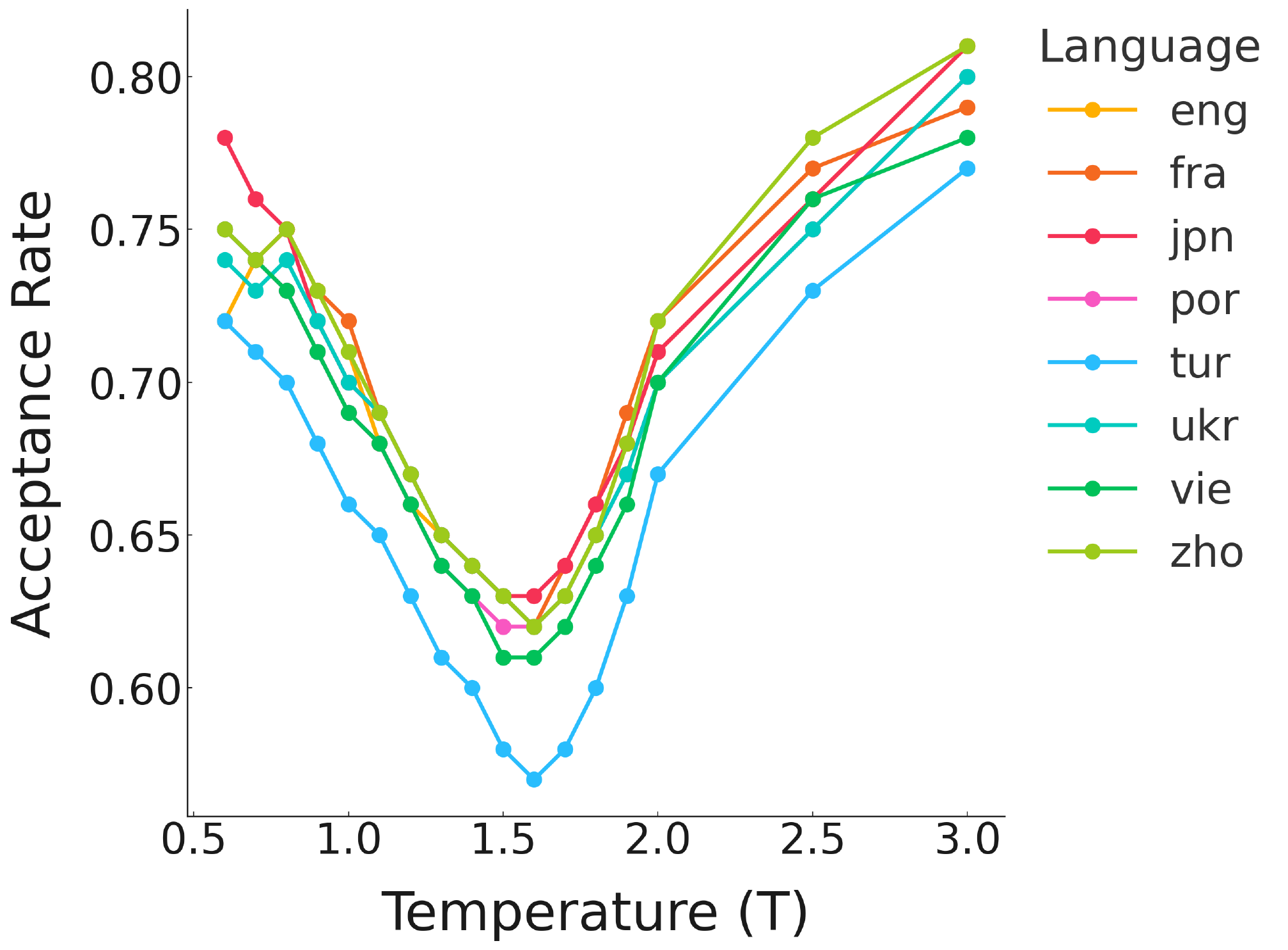} 
    \caption{Alpha at different temperatures for each language}
    \label{fig:7a}
  \end{subfigure} 
  \hspace{5pt}
  \begin{subfigure}[t]{0.44\textwidth}
    \centering
    \includegraphics[width=\linewidth,]{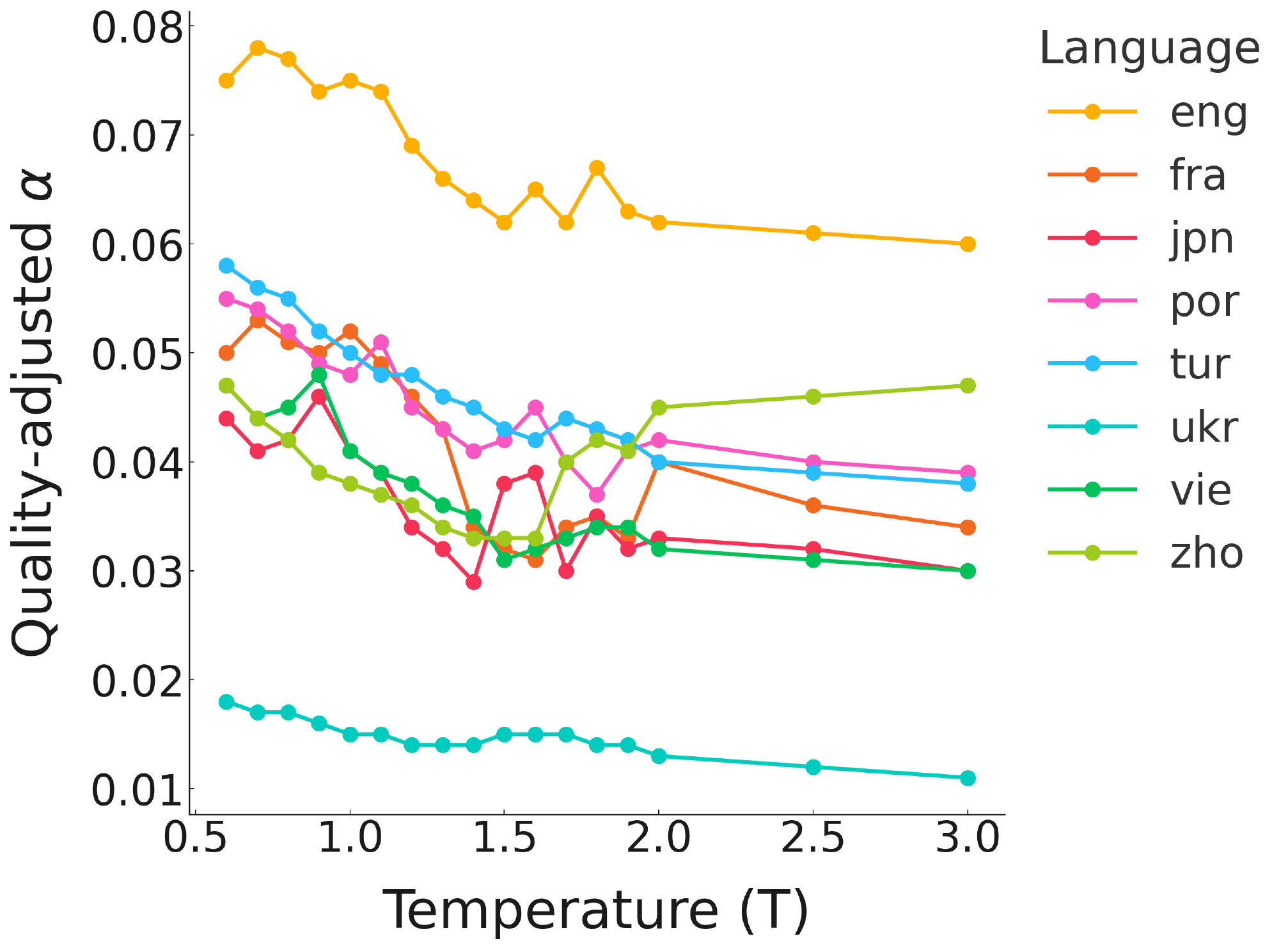} 
    \caption{Quality adjusted alpha at different temperatures for each language.}
    \label{fig:7b}
  \end{subfigure}
  \caption{Degrading influence of temperature over DollyQA data. Consistent parabolic alpha distributions, roughly flat quality-adjusted alpha distributions.}
  \label{fig:7}
  \vspace{-5pt}
\end{figure}

\textbf{Joint temperature optimization.}
When verifier and drafter temperatures move in tandem, higher temperatures bring distributions toward uniformity, increasing speed-up but degrading generation quality. Similarly, one-hot distributions at temperature zero may lead to higher acceptances, however, this set-up suffers from the same quality degradation, especially in smaller verifiers \citep{nakaishi2024criticalphasetransitionlarge}. Experiments on DollyQA show that $\alpha$ thus follows a parabolic trend over temperature, with minima around $T \approx 1.5$ (Figure~\ref{fig:7a}). However, quality-adjusted speed-up—computed as $\alpha \cdot \beta$, where $\beta$ is BLEU—remains somewhat flat across temperature settings (Figure~\ref{fig:7b}). This suggests that while temperature affects $\alpha$, it does so in a way that undermines output quality, offering no practical fairness benefit. We accordingly focus our attention to methods that do not influence verifier generations.

\begin{figure}[t]
  \centering
  \begin{subfigure}[t]{0.32\textwidth}
    \centering
    \includegraphics[width=\linewidth]{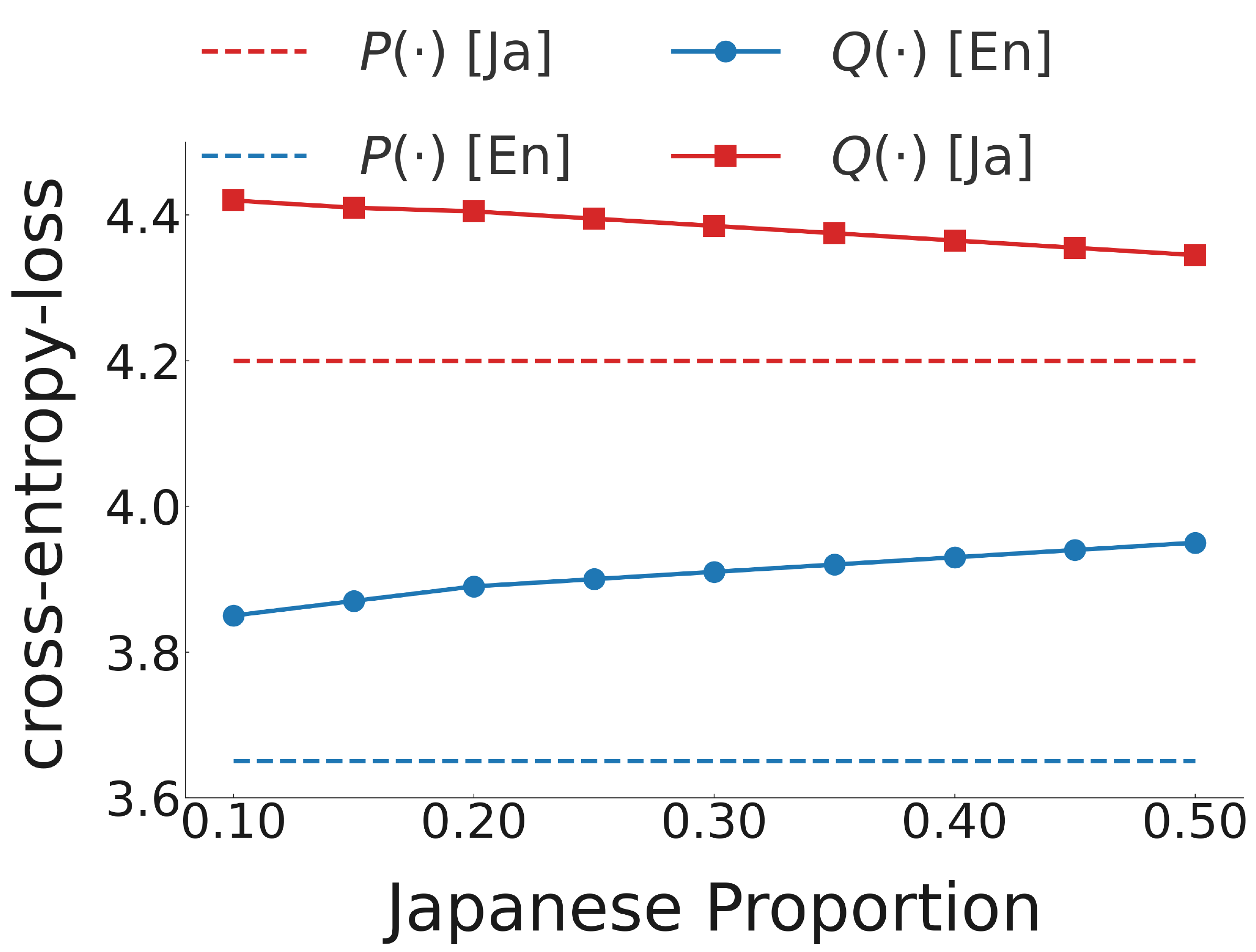}
    \caption{$P$ and $Q$ losses on `En` and `Ja` during progressive finetuning.}
    \label{fig:lossbi}
  \end{subfigure}\hfill
  \begin{subfigure}[t]{0.32\textwidth}
    \centering
    \includegraphics[width=\linewidth]{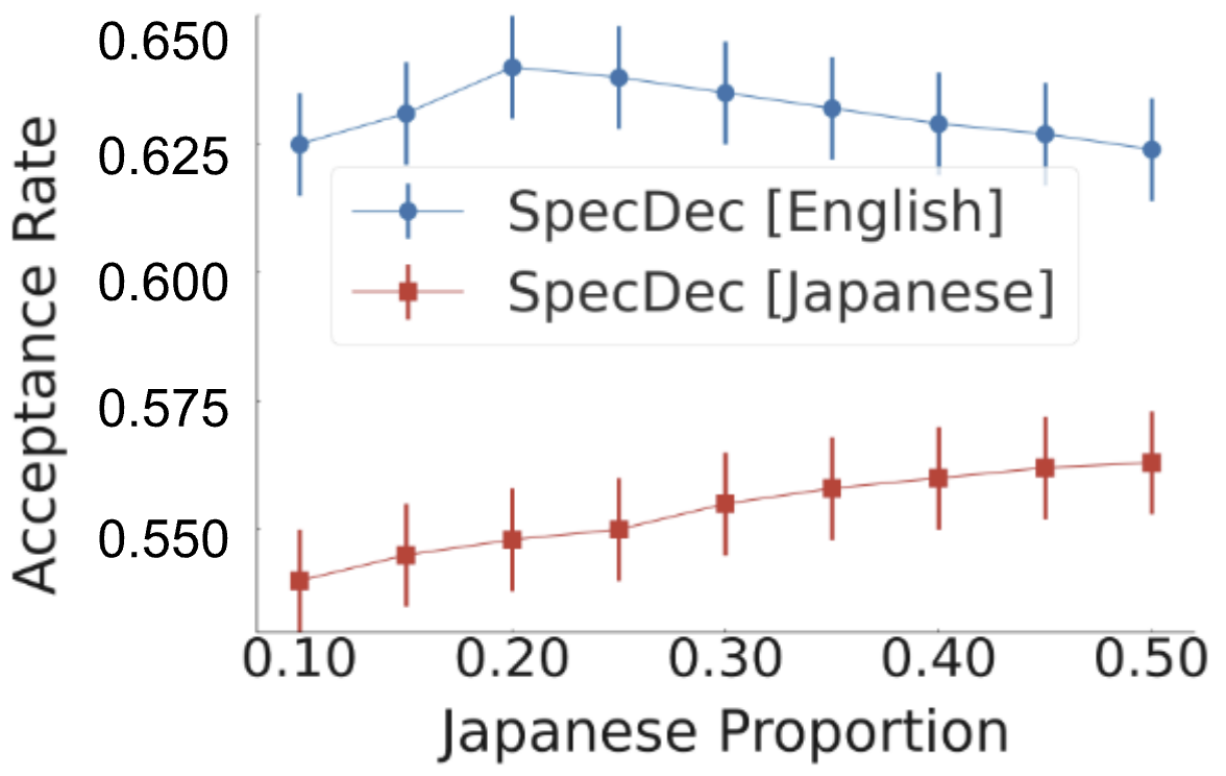}
    \caption{Language acceptance rates during data progressive finetuning.}
    \label{fig:abi}
  \end{subfigure}\hfill
  \begin{subfigure}[t]{0.32\textwidth}
    \centering
    \includegraphics[width=\linewidth]{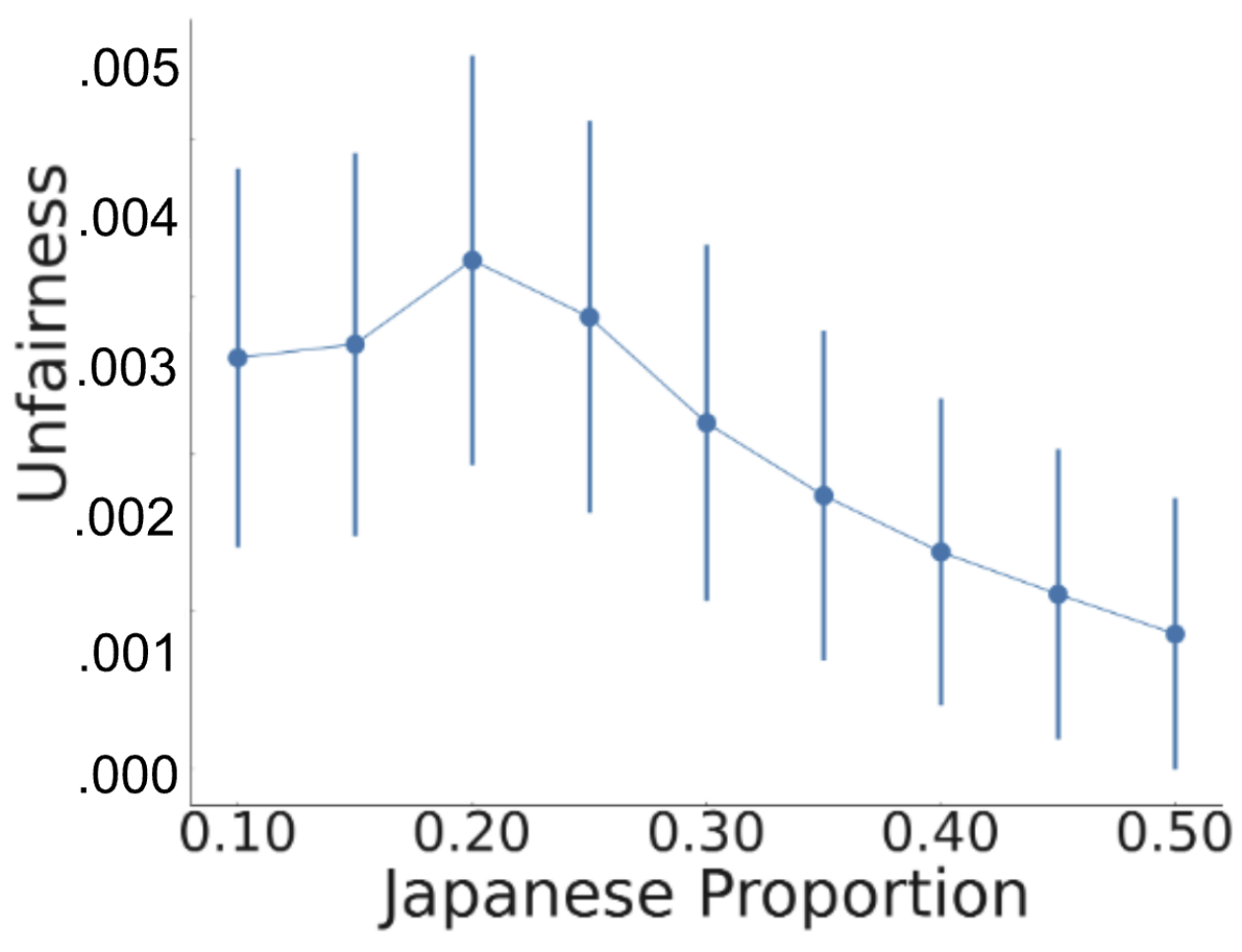}
    \caption{Unfairness objective during data progressive finetuning.}
    \label{fig:ubi}
  \end{subfigure}          
  \caption{Data progressive finetuning in bilingual setting.}
  \label{fig:three-wide}
\end{figure}

\textbf{Data balancing.}
We next investigate the influence that differing data proportions during finetuning have on speed-up fairness,  exploring a mixture-based finetuning strategy in the bilingual (English-Japanese) setting. Varying the proportion of Japanese data shows that increasing its prevalence improves Japanese drafter loss and acceptance rates, while inducing minor regressions in English performance (Figures \ref{fig:lossbi}, \ref{fig:abi}).
Unfairness, ${\cal U}(\cdot)$, decreases as Japanese data increases, showing that basic data re-balancing can have beneficial effects, as well as shows the effects data representation has on speed-ups (Figure \ref{fig:ubi}). However, this approach lacks principled guidance for setting data ratios, and is not clearly scalable to several tasks, prompting further exploration.

\textbf{Stochastic corrective drafter finetuning (s-CDF)}. Recall our proposed unfairness function $\cal U$. We utilize our derived projected gradient, Equation \ref{eq:proj_grad_constr}, to take descent steps along our function, $\cal U$. We find that for a set of languages, optimizing the objective $\cal U$ stochastically, is the most practical, while still leading to unfairness convergence. In these experiments, we use a batch size of $\mathcal B =8$, and select five languages from our MCoT dataset that show initial unfairness (English, Spanish, Russian, Mandarin, German). We repeat experiments over multiple model pairs from the Qwen2.5 series, testing across five model pairs with 0.5B, and 1.5B parameter models as drafters, and 3B, 7B, and 14B parameter models as verifiers. We see on average, a \emph{20\%} reduction in the variance of our acceptance rates across our model pairs during finetuning, alongside, a \emph{12\%} decrease in our unfairness $\cal U$. The mutual reduction in speed-up variance and speed-up unfairness $\cal U$ serves to showcase the connection between our unfairness metric and speed-up dispersions empirically, as was established in Theorem~\ref{thm:chain}.



\begin{figure}[t]
  \centering
  \begin{subfigure}[t]{0.50\textwidth}
    \centering
    \includegraphics[width=\linewidth]{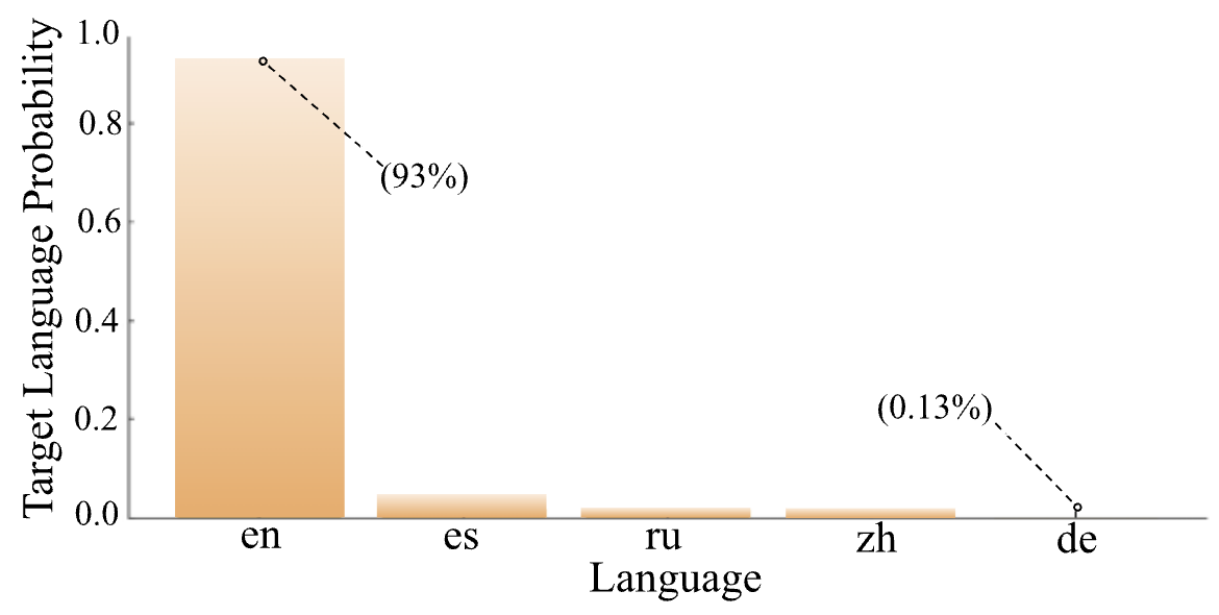}
    \vspace{-10pt}
    \caption{Target language probability for each language during finetuning, with 'true' target language (en) highlighted. \hspace{8pt}}
    \label{fig:rstar}
    \vspace{-35pt}
  \end{subfigure}
  \vspace{12pt}
  \hspace{12pt}
  \begin{subfigure}[t]{0.40\textwidth}
    \centering
    \includegraphics[width=\linewidth]{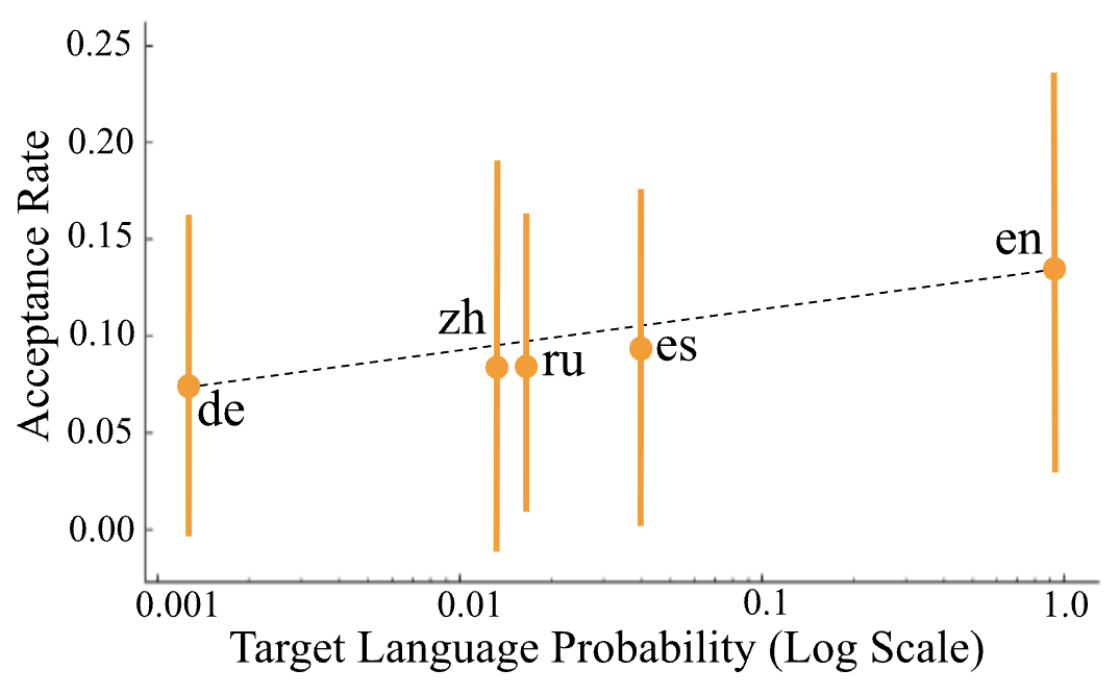}
    \caption{Acceptance rate correlation with target probability}
    \label{fig:rstar_alpha}
  \end{subfigure}\vspace{-8pt}
  
  \caption{Language sampling probabilities and acceptance rates during the proposed corrective drafter tuning (s-CDF)}
\end{figure}

We also see that our fastest language, English, is sampled as our target language in 93\% of cases, Figure \ref{fig:rstar}, in other words giving English a 'target language probability' of 93\%, while our slowest language, German, features gets sampled as $\bm{D}_{\min}$ ~0.13\% of batches, Figure \ref{fig:rstar}. This showcases that our stochastic approach tends to preserve information about speed-up disparities even with small batch sizes ($\mathcal B = 8$), as well as speaks to the persistent speed-up unfairness present in multilingual datasets. Subsequently, when studying the relationship between the target language probability and the acceptance rates, we observe a positive trend between the target probability for a language, and the acceptance rates for the language $\alpha$, see \ref{fig:rstar_alpha}, showcasing the predictive power of our divergence metric $\bm{D}(\cdot)$ at forecasting speed-ups even in stochastic contexts.

\section{Conclusion}
\label{sec:conclusion}


This work reveals a previously overlooked source of unfairness in accelerated inference: speculative decoding yields unequal speed-up benefits across tasks and languages. We find that underrepresented or under-fit distributions—such as low-resource languages consistently receive lower speed-up motivated by disparities in drafter fitness. To understand and mitigate this disparity, we conducted a comprehensive empirical study across multilingual benchmarks, and show the consistency of this fairness issue. This analysis was driven by theoretical intuitions presented in our work, showing a connection between task-wise acceptance and the drafter task fitness. 
Finally, we proposed a mitigation technique based on a projected gradient-based method that reduces speed-up disparities by selectively improving under-performing tasks.
We believe these results are important as they draw attention to a potential source of computational inequity and that guaranteeing both accuracy parity and acceleration parity across populations is an area deserving attention.


\newpage
\section*{Ethics Statement}

Our work concerns the ethical deployment of speculative decoding in a manner that is inherently fairness-aware with respect to various, potentially discriminated subgroups. Our analysis cautions practitioners against ignoring the disparate effects that inference acceleration algorithms can have on certain subgroups. However, the warnings communicated in this paper, strictly speaking, do not prevent the misuse or irresponsible usage of speculative decoding yet the emphasis on mitigation techniques, we believe, contributes to the open problem of fair, responsible and ethical AI usage.

\section*{Reproducibility Statement}

We documented all artifacts required to reproduce our results in the main paper and Appendix: checkpoints for speculative decoding (draft/verify), our fairness metric $\mathcal{U}$ (cross-entropy–based), logging utilities (acceptance rate, tokens-per-step, realized speed-up), and s-CDF algorithms. Upon release, our repository includes exact \emph{model identifiers}, tokenizer versions, acceptance criteria, and hyperparameters; configuration files (YAML) enumerate drafter–verifier pairs, batch sizes, maximum lengths, and stopping rules. We fix and document \emph{random seeds} (for data sampling and any stochastic training), report \emph{hardware} (GPU model, driver/CUDA versions) and key libraries used, and provide command lines to reproduce every table/figure. For datasets, we provide citation instructions and will publish \emph{frozen evaluation lists} (document IDs and prompts) to avoid data drift. 
Finally, we also plan to report wall-clock compute and carbon estimates for major runs. 

\section*{Acknowledgments}
This research was partially funded by NSF awards RI-2533631, SaTC-2133169, RI-2232054, and CAREER-2143706, and a Fellowship in AI Research from the LaCross Institute. The authors also acknowledge the Research Computing at the University of Virginia. The views and conclusions of this work are those of the authors only.

\bibliography{custom}

\newpage

\appendix
\section{Missing Proofs}
\label{A:proofs}

\setcounter{theorem}{0}

\subsection{Relating Divergence and Speed-up via a Monotone Chain }\label{A:tm1}
\begin{theorem}
For $\gamma\ge 1$, $f_\gamma:[0,1)\to\mathbb{R}_+$ is strictly increasing (and convex for $\gamma\ge2$). Consequently, for fixed $(\gamma,c)$, there is a monotone chain from $S_T$ to $\bm{D}_T$:

\begin{equation}
        S_T \;\ge\; \frac{f_\gamma\!\Big(1-\sqrt{\tfrac{1}{2}\,\mathrm{KL}_T}\Big)}{\gamma c+1}
        \;\ge\;
        \frac{f_\gamma\!\Big(1-\sqrt{\tfrac{1}{2}\,\bm{D}_T}\Big)}{\gamma c+1}\,,
\end{equation}

\emph{where $T$ is a task-distribution over prefixes, $f_\gamma(\alpha)=\frac{1-\alpha^{\gamma+1}}{1-\alpha}$ is expected accepted tokens, and $S_T$ is the task-wise speed-up, defined as $S_T=\;\frac{f_\gamma(\alpha(s))}{\gamma c+1}.$ Finally, task-wise Kullback–Leibler and cross-entropy are $\,\mathrm{KL}_T=\mathbb{E}_{s\sim T}[\mathrm{KL}(p\Vert q)]\,$ and $\,\bm{D}_T=\mathbb{E}_{s\sim T}[H(p,q)]$ with $H(p,q)=\mathbb{E}_{x\sim p}[-\log q(x)]$, for verifier, drafter posteriors $p(x), q(x)$ respectively on next-token $x$.}

\vspace{5pt}

\begin{proof}
    
First, $f_\gamma'(\alpha)=\sum_{k=1}^{\gamma}k\,\alpha^{k-1}>0$ for $\alpha\in[0,1)$, so $f_\gamma$ is strictly increasing; and $f_\gamma''(\alpha)=\sum_{k=2}^{\gamma}k(k-1)\alpha^{k-2}\ge0$, with strict convexity for $\gamma\ge2$ on $(0,1)$. 

By Jensen on convex $f_\gamma$ ($\gamma\ge2$; equality when $\gamma=1$ as $f_\gamma$ is affine): 
\[
\mathbb{E}[f_\gamma(\alpha(s))] \;\ge\; f_\gamma(\mathbb{E}[\alpha(s)]) \;=\; f_\gamma(\alpha_T).
\]
Divide by $\gamma c+1$ to get the bound for $S_T$.

By Pinsker, $\mathrm{TV}(p,q)\le\sqrt{\tfrac{1}{2}\mathrm{KL}(p\Vert q)}$, hence $\alpha(s)=1-\mathrm{TV}(p,q)\ge 1-\sqrt{\tfrac{1}{2}\mathrm{KL}(p\Vert q)}$. Taking expectations and concavity of the square root yields 
$\alpha_T\ge 1-\sqrt{\tfrac{1}{2}\,\mathrm{KL}_T}$. Using $D_T=H(p)+\mathrm{KL}_T\ge\mathrm{KL}_T$ implies the second inequality. Monotonicity of $f_\gamma$ finishes the chain. 
\end{proof}

\end{theorem}

\subsection{Speed-up and Acceptance Dependency} \label{A:corr1} 
\begin{corollary} For $\gamma\ge 1$, $f_\gamma:[0,1)\to\mathbb{R}_+$ strictly increasing, we conclude that $\,\mathrm{Speedup}(s;\gamma,c)$ is strictly increasing in $\alpha(s)$.

\begin{proof}
    $f_\gamma'(\alpha)=\sum_{k=1}^{\gamma}k\,\alpha^{k-1}>0$ for $\alpha\in[0,1)$, so $f_\gamma$ is strictly increasing, as stated, with strict convexity for $\gamma\ge2$ on $(0,1)$. Therefore the claim follows since $\mathrm{Speedup}(s;\gamma,c)=\tfrac{f_\gamma(\alpha(s))}{\gamma c+1}$ shares the monotonicity of $f_\gamma$.
\end{proof}
\end{corollary}

\subsection{Relating Drafter-Fitness and Acceptance} \label{A:thm2}
\begin{theorem}

Let $u(\cdot\mid s)$ be the latent task posterior. Define task misfits
\[
r_p\;\triangleq\;\mathbb{E}_{s\sim T}\Big[\tfrac{1}{2}\!\sum_{x\in V}\!\big|u(x\mid s)-p(x\mid s)\big|\Big],\qquad
r_q\;\triangleq\;\mathbb{E}_{s\sim T}\Big[\tfrac{1}{2}\!\sum_{x\in V}\!\big|u(x\mid s)-q(x\mid s)\big|\Big].
\]
Assume $r_p\le r_q$ (typical in practice). Then
\[
\big|\alpha_T-(1-r_q)\big|\;\le\; r_p.
\]

\begin{proof}
    By triangle inequality and its reverse for total variation at each prefix $s$:
    \[
    \big|\,\mathrm{TV}(p,q)-\mathrm{TV}(u,q)\,\big| \;\le\; \mathrm{TV}(u,p) \;\Rightarrow\;
    \big|(1-\alpha(s))-(r_q(s))\big|\;\le\; r_p(s),
    \]
    where $r_q(s)=\mathrm{TV}(u,q)$ and $r_p(s)=\mathrm{TV}(u,p)$. Averaging over $s\sim T$ and using Jensen on $\big|(1-\alpha(s))-(r_q(s))\big|\;\le\; r_p(s)$ yields $\big|(1-\alpha_T)-r_q\big|\le r_p$, i.e., $\big|\alpha_T-(1-r_q)\big|\le r_p$.
    \end{proof}
\end{theorem}

\subsection{A sufficient Condition For Disparities} \label{A:corr2}
\begin{corollary}

For two tasks $T_i,T_j$ with pairs $(r_p^i,r_q^i)$ and $(r_p^j,r_q^j)$, a strict acceptance gap $\alpha_{T_i}>\alpha_{T_j}$ is guaranteed whenever
\[
r_q^j - r_q^i \;>\; r_p^i + r_p^j.
\]
Moreover, by Corollary~1.,
\[
S_{T_i} - S_{T_j} \;\ge\; \frac{\alpha_{T_i}-\alpha_{T_j}}{\gamma c+1}\;>\;0.
\]
\begin{proof}
From Theorem~2, $\alpha_{T_i}\ge 1-(r_q^i+r_p^i)$ and $\alpha_{T_j}\le 1-(r_q^j-r_p^j)$ (since $r_q^j\ge r_p^j$ by assumption). The stated condition implies $1-(r_q^i+r_p^i) > 1-(r_q^j-r_p^j)$, hence $\alpha_{T_i}>\alpha_{T_j}$. Apply Corollary~1. 
\end{proof}
\end{corollary}

\subsection{Correctness Of Speculative Sampling} \label{A:thm3}
\begin{theorem}
Tokens sampled via \emph{speculative sampling} from $p(x)$ and $q(x)$ are distributed identically to those sampled from $p(x)$. 

\begin{proof}
(As reported in \citet{Leviathan_2023}) Let $\alpha$ be the acceptance probability. Note that as $p'(x) = \operatorname{norm}(\max(0, p(x) - q(x))) = \frac{p(x) - \min(q(x), p(x))}{\sum_{x'}(p(x') - \min(q(x'), p(x')))} = \frac{p(x) - \min(q(x), p(x))}{1 - \alpha}$, the normalizing constant for the adjusted distribution $p'(x)$ is $1 - \alpha$.

Now:

$$
    P(x=x') = P(\text{guess}\ \text{accepted}, x=x') + P(\text{guess}\ \text{rejected}, x=x')
$$
Where:
\begin{align}
    P(\text{guess}&\ \text{accepted}, x=x') = q(x')\min(1, \frac{p(x')}{q(x')}) = \min(q(x'), p(x'))\\
    P(\text{guess}&\ \text{rejected}, x=x') = (1 - \alpha)p'(x')  = p(x') - \min(q(x'), p(x'))
\end{align}
And thus, overall we obtained the sought result:
\begin{equation}
    P(x=x') = \min(p(x'), q(x')) + p(x') - \min(p(x'), q(x')) = p(x').
\end{equation}
\end{proof}
\end{theorem}

\section{Extended Related Work}\label{A:related_work}

\paragraph{Fairness in LLMs.}
With the widespread use of \glspl*{llm}, concerns regarding fairness have become increasingly prominent. Notably, fairness-related issues in \glspl*{llm} can manifest in the form of toxicity in generated outputs and biases that cause harm to various social groups. In particular, a well-established distinction \citep{ChuWZ2024SIGKDD,GallegosRBTKDYZA2024ACL} exists between \emph{representational harms}—such as the use of derogatory language, disparate system performance, erasure, misrepresentation, and stereotyping, which contribute to denigrating and subordinating attitudes toward certain social groups—and \emph{allocational harms}, which involve the amplification of existing biases, the creation of new biases, and the reinforcement of stereotypes.

While significant efforts have been made to address these issues through various fairness metrics \citep{DhamalaSKKPCG2021FaccT,DelobelleTCB2022NAACL,CzarnowskaVS2021TACL} and mitigation techniques \citep{SteedPKW2022ACL,OmraniZYGKAJD2023ACL,KumarLZCESR2023EACL}, the problem remains far from solved, as biases can be introduced at multiple stages, including through training prompts, labeling choices, or at the embedding level.

For a comprehensive overview of fairness issues in LLMs, we refer the reader to the survey by \citet{GallegosRBTKDYZA2024ACL} and the taxonomic survey by \citet{ChuWZ2024SIGKDD}.

\paragraph{Multilingual LLMs.}
In this work, we focus on the issue of disparate system performance across different populations, where these populations are represented by different languages. Specifically, we examine fairness across languages in the context of \emph{speculative decoding}, where disparities may arise due to varying time and cost requirements for generating outputs in different languages.

The problem of fairness in multilingual \glspl*{llm} has been studied from various perspectives. For instance, \citet{PetrovMTB2023NeurIPS} highlight how language model tokenizers introduce unfairness between languages, while \citet{YiTHDSY2024EMNLP} demonstrate that language-specific draft models, optimized through a targeted pretrain-and-finetune strategy, significantly improve inference speed compared to previous methods.

Importantly, while the study of the impact of techniques to speed up and reduce the impact of using \glspl*{llm} has already been investigated \citep{DasRTRKF2024CoRR}, our study combines speculative decoding fairness and multilingual \glspl*{llm}s. In addition, we opt to offer detailed explanation for these effects and rigorous mitigation strategies in a manner that is distinct from previous works.


\paragraph{Speculative decoding.}
Where prior works like \citet{yi2024fastmultilingualllminference} examine the deployment of speculative decoding in multilingual settings, evaluating over multiple sub-tasks, they ignore critical fairness questions which we address. Additionally, works like \citet{Zhou_2023_DistillSpec} discuss different notions of distributional misalignment, which they use to optimize speed-up on tasks. However, they do not evaluate speed-up optimization in the context of multiple sub-tasks, and do not extend their analysis to evaluate the relationship between task fitness and drafter, verifier divergence.

\section{Additional Unfairness Evidence} \label{A:unfairness}

\begin{figure}[h]
  \centering
\includegraphics[width=0.7\linewidth]{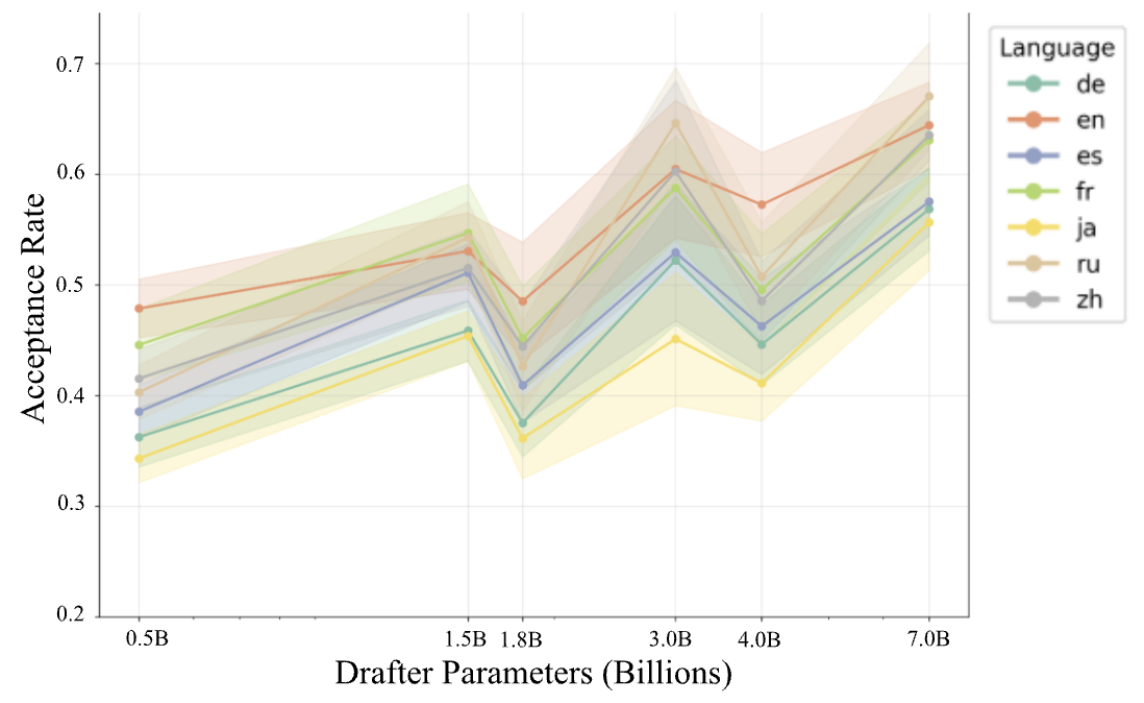}
  \caption{MGSM data, Qwen2.5 verifier (3B), various drafters: [Qwen2.5-(0.5B, 1.5B, 3B-base), and  Qwen1.5-(4B, 7B)]. Acceptance scaling properties with drafter parameters. Speed-up hierarchy shows consistency across drafter models regardless of scale.}
  \label{fig:draft_abl_mgsm}
\end{figure}

We highlight, in Figure \ref{fig:draft_abl_mgsm} that MGSM speed-up disparities persist across different drafter models. If we fix the verifier, Qwen2.5-3B, and ablate over drafter models we observe that slow languages are consistently slow, and fast languages are consistently fast. Notably Japanese is the slowest language across all tested drafters, and English is the fastest language in 50\% of cases. This ablation further speaks to the consistency of multilingual speed-up disparities.

\section{Extended Discussion And Reproducibility Details}
\label{sec:extended-results}

\subsection{s-CDF: Training Procedure}
We implement s-CDF as a light-touch finetuning loop over languages (tasks) that (i) uses \emph{teacher-forced completions from the verifier} as supervision for the drafter, and (ii) \emph{scales gradients per-language} by their excess misfit over the current best language $r^\star$ to minimize variance from the minimum.\footnote{Intuition: approximate the projected gradient of $\sum_T (D_T-D_{\min})^2$ where $D_T$ is task cross-entropy; this retains a certificate that increasing fitness raises a lower bound on $\alpha_T$ and therefore $S_T$.}

\paragraph{Models and quantization.}
We load a student/drafter $q$ and teacher/verifier $p$ with 4-bit NF4 quantization (bitsandbytes) and bfloat16 compute; the drafter enables gradient checkpointing and disables \texttt{use\_cache} to reduce memory during training, while the verifier enables \texttt{use\_cache} for fast generation.

\paragraph{Data pipeline.}
We read a JSON of records with a \texttt{lang} field and question text; we index records per-language and tokenize prompts to a fixed \texttt{MAX\_PROMPT\_TOK}. Labels are retained on prompt tokens (non-padding) to compute a prompt-TV proxy later. 

\paragraph{Teacher completions and student loss.}
Given a mini-batch of prompts $X$, the teacher $p$ generates up to \texttt{MAX\_GEN\_TOKENS} tokens (sampling) with cache enabled; we splice out $S$ (new tokens) and form \texttt{seq\_all = [X; S]} with a full-attention mask. The student $q$ is run on \texttt{seq\_all} and trained with cross-entropy on the $S$ segment only (no prompt loss). This aligns $q$ to $p$'s next-token distribution on continuation tokens.

\paragraph{Per-language misfit and gradient shaping.}
For each language $\ell$ sampled this step we compute $r_\ell=\text{CE}(q\Vert p)$ (mean over the mini-batches of that language). We identify the current best language $\ell^\star=\arg\min_\ell r_\ell$ with value $r^\star$ and scale each $r_\ell$'s gradient by $(r_\ell-r^\star)/(\texttt{GRAD\_ACCUM}\cdot |\{\ell\}|)$ before updating $q$ (gradient clipping and accumulation supported). This implements the \emph{shift-by-minimum} s-CDF update. 

\paragraph{Acceptance and prompt-TV logging.}
At user-configured cadence we estimate acceptance rate with a \emph{single-pass} proxy: draft $\gamma$ tokens with $q$ (no sampling), score the same positions with $p$, accept if $\min\{1,p/q\}$ exceeds a uniform random draw per token, and aggregate the \emph{contiguous} accepted prefix length per row normalized by total drafted tokens. This tracks how alignment changes translate into realized acceptance. We also log a prompt-TV proxy for $p$ and $q$ (\,$1-\Pr[\text{correct token}]$ on prompt positions\,). Metrics are CSV-logged with timestamps. 

\paragraph{Optimizers and stability.}
When quantized parameters are present we favor \texttt{AdamW8bit} (bitsandbytes), otherwise standard \texttt{AdamW}. We clip gradients to \texttt{CLIP} before stepping every \texttt{GRAD\_ACCUM} mini-batches.

\paragraph{Default hyperparameters (reproducible starting point).}
Unless stated, we used: \texttt{STEPS}=10{,}00; \texttt{SAMPLE\_LANGS\_PER\_STEP}=5; \texttt{BATCH\_PER\_LANG}=64 prompts; \texttt{MINI\_BATCH\_SIZE}=8; \texttt{MAX\_PROMPT\_TOK}=512; \texttt{MAX\_GEN\_TOKENS}=64; \texttt{LR}=1e-4; \texttt{GRAD\_ACCUM}=4; \texttt{CLIP}=1.0; acceptance draft width $\gamma=5$ in the estimator.

\subsection{Experimental Setup}
\textbf{Hardware.} Training/ablation runs were conducted on single-GPU nodes (e.g., 1$\times$A6000~48GB). The script constrains \texttt{max\_memory} and uses 4-bit NF4 quantization for both $q$ and $p$ to fit comfortably.

\textbf{Software.} PyTorch (bf16 compute), HuggingFace \texttt{transformers}, bitsandbytes for quantization/optimizer; gradient checkpointing enabled on the drafter to control memory; \texttt{use\_cache} toggled as described. 

\textbf{Datasets and splits.} Following the paper, we evaluate multilingual math reasoning (MGSM, MCoT) and general instruction following (Dolly/Aya subset), reporting per-language acceptance/speed-ups and task metrics. See results sections and Appendix \ref{A:unfairness} for extended evidence of persistent disparities across drafters and datasets. 

\subsection{Main Results}
\paragraph{Multilingual speed-up disparities persist without mitigation.}
Consistent with Section \ref{sec:motivation}, acceptance and accuracy vary significantly by language; low-accuracy languages exhibit the lowest acceptance and thus the smallest speed-ups. We observe a persistent hierarchy across drafter scales (Qwen2.5 0.5–14B) with Japanese repeatedly among the slowest and English always the fastest, and the ordering is stable when swapping datasets (MGSM (Small) $\rightarrow$ MGSM (MCoT)). 

\paragraph{Throughput vs.\ quality.}
Given that s-CDF never updates the verifier, the target distribution remains unchanged; by construction the drafter becomes a better proposal distribution w.r.t.\ $p$, and acceptance increases without altering $p$'s vanilla behavior. Quality metrics under vanilla decoding with $p$ remain stable; drafter-only generations improve on continuations seen during training due to student-on-teacher learning, but we do not use drafter-only decoding at inference time.

\subsection{Reproducibility Checklist}
\begin{itemize}
\item \textbf{Data:} provide a JSON with fields \texttt{lang}, \texttt{question}; ensure each selected language has at least one record or it will be dropped.
\item \textbf{Batching:} choose \texttt{SAMPLE\_LANGS\_PER\_STEP}, \texttt{BATCH\_PER\_LANG}, \texttt{MINI\_BATCH\_SIZE} to fit memory; gradient-accumulate with \texttt{GRAD\_ACCUM}. 
\item \textbf{Tokenization limits:} set \texttt{MAX\_PROMPT\_TOK}, \texttt{MAX\_GEN\_TOKENS}; pad-token is set to EOS if missing.
\item \textbf{Optimizer:} use AdamW8bit when quantized, else AdamW; clip to \texttt{CLIP}. 
\item \textbf{Logging:} CSV columns include timestamp, step, \texttt{star\_lang}, \texttt{lang}, $r_\ell$, acceptance, prompt-TV for $q$ and $p$.
\item \textbf{Acceptance estimator:} keep default $\gamma{=}5$ (configurable); periodicity via \texttt{EVAL\_EVERY}. 
\end{itemize}

\subsection{Limitations and Practical Notes}

\textbf{Estimator bias.} The logged acceptance proxy uses greedy drafts and per-token $\min\{1,p/q\}$ tests; it underestimates streak acceptance when deployment uses different temperatures or sampling filters. Use it for trend-tracking, not absolute certification. \newline
\textbf{Data coverage.} When languages have very few prompts, $r^\star$ can be noisy; we recommend deploying in scenarios with large datasets.
\textbf{Compute/latency.} Realized throughput depends on packing/verifier parallelism; speed-up equations assume negligible overhead when verifying $\gamma$-drafted tokens in one pass. Validate on your specific hardware.

However, note that s-CDF is quite simple to implement (by design), works with 4-bit quantized $q$/$p$, and \emph{directly} targets the acceptance gap that governs speculative speed-ups.

\end{document}